\documentclass[letterpaper,journal]{IEEEtran}
\usepackage{amsmath,amsfonts}
\usepackage{algorithmic}
\usepackage{algorithm}
\usepackage{array}
\usepackage[caption=false,font=normalsize,labelfont=sf,textfont=sf]{subfig}
\usepackage{textcomp}
\usepackage{stfloats}
\usepackage{url}
\usepackage{verbatim}
\usepackage{graphicx}
\usepackage{cite}
\usepackage{multirow}
\usepackage{color}
\usepackage{booktabs}
\usepackage[table]{xcolor}
\usepackage{colortbl}
\usepackage{pifont}
\usepackage{marvosym}
\usepackage{animate}
\usepackage{hyperref}
\renewcommand{\thesection}{\arabic{section}}
\renewcommand{\thesubsection}{\thesection.\arabic{subsection}}
\renewcommand{\thesubsubsection}{\thesubsection.\arabic{subsubsection}}

\makeatletter
\def\@seccntformat#1{\csname the#1\endcsname.\quad} %
\makeatother

\begin{document}

\newcommand{\NR}{--}
\newcommand{\redx}{\textcolor{red}{\ding{55}}}

\definecolor{commentcolor}{RGB}{110,154,155}   %
\newcommand{\PyComment}[1]{\ttfamily\textcolor{commentcolor}{\# #1}}  %
\newcommand{\PyCode}[1]{\ttfamily\textcolor{black}{#1}} %

\newcommand{\methodname}{\textup{\textsc{Neurons++}}}
\newcommand{\methodnamebf}{\textbf{\textup{Neurons++}}}
\newcommand{\confname}{\textup{\textsc{Neurons}}}

\title{\methodname{}: A Unified Visual-Stream Framework for Image and Video Reconstruction from fMRI with Neural-to-Visual Interpretability}

\author{Haonan Wang, Yuchen Wang, Jingyu Lu, Qixiang Zhang, Lehan Wang, Xuanqi Huang, and Xiaomeng Li \thanks{Haonan Wang, Yuchen Wang, Jingyu Lu, Qixiang Zhang, Lehan Wang, Xuanqi Huang, and Xiaomeng Li are with The Hong Kong University of Science and Technology. Yuchen Wang, Jingyu Lu and Xiaomeng Li are also with Shenzhen Loop Area Institute, Shenzhen, China.} \thanks{Corresponding author: Xiaomeng Li. E-mail: eexmli@ust.hk. } }

\maketitle
\thispagestyle{plain}

\begin{abstract}
Decoding visual experience from fMRI is important for understanding human perception and developing brain-computer interfaces. Recent generative decoding methods can reconstruct plausible images and videos, but visual plausibility alone does not ensure that the output is faithfully grounded in brain activity. In this work, we propose \methodname{}, a visual-stream fMRI reconstruction framework for both static image and dynamic video reconstruction. Compared with our conference version \textsc{Neurons}, \methodname{} extends the paradigm from video-only reconstruction to visual-stream reconstruction for both images and videos, and replaces parallel task conditioning with serial cortex-inspired decoding. The model first uses image-fMRI pretraining to obtain a stable spatial-semantic initialization, and then performs subject-specific fine-tuning for image and video reconstruction separately. The serial design organizes fMRI-derived representations into explicit spatial, semantic, language-level, and temporal cues. Extensive experiments show that \methodname{} achieves state-of-the-art video reconstruction and competitive image reconstruction across standard benchmarks. Beyond reconstruction fidelity, neural-to-visual analyses indicate that \methodname{} exhibits qualitative directional controllability under neural perturbation, that its fMRI encoder read-out is sensitive to stimulus-relevant cortical features, and that it supports ROI-guided semantic modulation. Code and model weights are available at: \href{https://github.com/xmed-lab/NeuronsPlusPlus}{\textit{\texttt{https://github.com/xmed-lab/NeuronsPlusPlus}}}.
\end{abstract}

\section{Introduction}

Decoding visual experience from human brain activity is a fundamental problem in cognitive neuroscience and brain-computer interfaces. Early studies have shown that fMRI responses contain rich information about perceived and imagined visual content, ranging from object categories to natural images and movies~\cite{horikawa2017firstfmri,nishimoto2011reconstructing,wen2018neural}. With the rapid development of vision-language models and generative priors, recent fMRI-based visual reconstruction methods have achieved increasingly realistic outputs. For static image reconstruction, modern methods typically map fMRI responses into CLIP~\cite{radford2021clip}, Stable Diffusion~\cite{rombach2022ldm}, or related latent spaces to synthesize natural images~\cite{takagi2023high,scotti2023reconstructing,mindeye2,gong2025mindtuner}. These methods demonstrate that large-scale visual priors can compensate for the sparsity and noise of fMRI signals and recover rich semantic and perceptual content from brain activity.

Despite this progress, three fundamental limitations remain.
First, existing methods are generally developed under \textbf{separated reconstruction settings}: image decoders emphasize spatial details and object semantics~\cite{takagi2023high,scotti2023reconstructing,mindeye2,xia2024umbrae,huo2024neuropictor,wang2024mindbridge,brainit,gong2025mindtuner}, whereas video decoders focus on motion and temporal continuity~\cite{wang2022reconstructing,kupershmidt2022penny,chen2024cinematic,gong2025neuroclips,wang2025neurons}.
This separation does not reflect continuous visual experience and prevents the relatively clean spatial-semantic supervision of image-fMRI data from supporting dynamic visual decoding.
Second, at the representation level, current methods also suffer from \textbf{disconnected neural-to-visual representations}.
Although recent approaches model low-level structure, high-level semantics, and temporal dynamics~\cite{brainit,yang2026semvideo,wang2025neurons}, these signals are often aligned implicitly or aggregated as independent conditioning cues rather than organized into a connected decoding hierarchy.
Consequently, it remains difficult to trace how different levels of neural information contribute to the reconstructed output.
Third, fMRI reconstruction still faces \textbf{unverified neural grounding}.
A visually plausible output may be dominated by generative priors rather than faithfully reflecting stimulus-relevant brain activity, especially when ground-truth stimuli are unavailable, as in imagined visual decoding or communication for patients with severe motor impairment~\cite{kneeland2025nsd_imagery}.

\begin{figure*}[t]
\centering
\includegraphics[width=\linewidth]{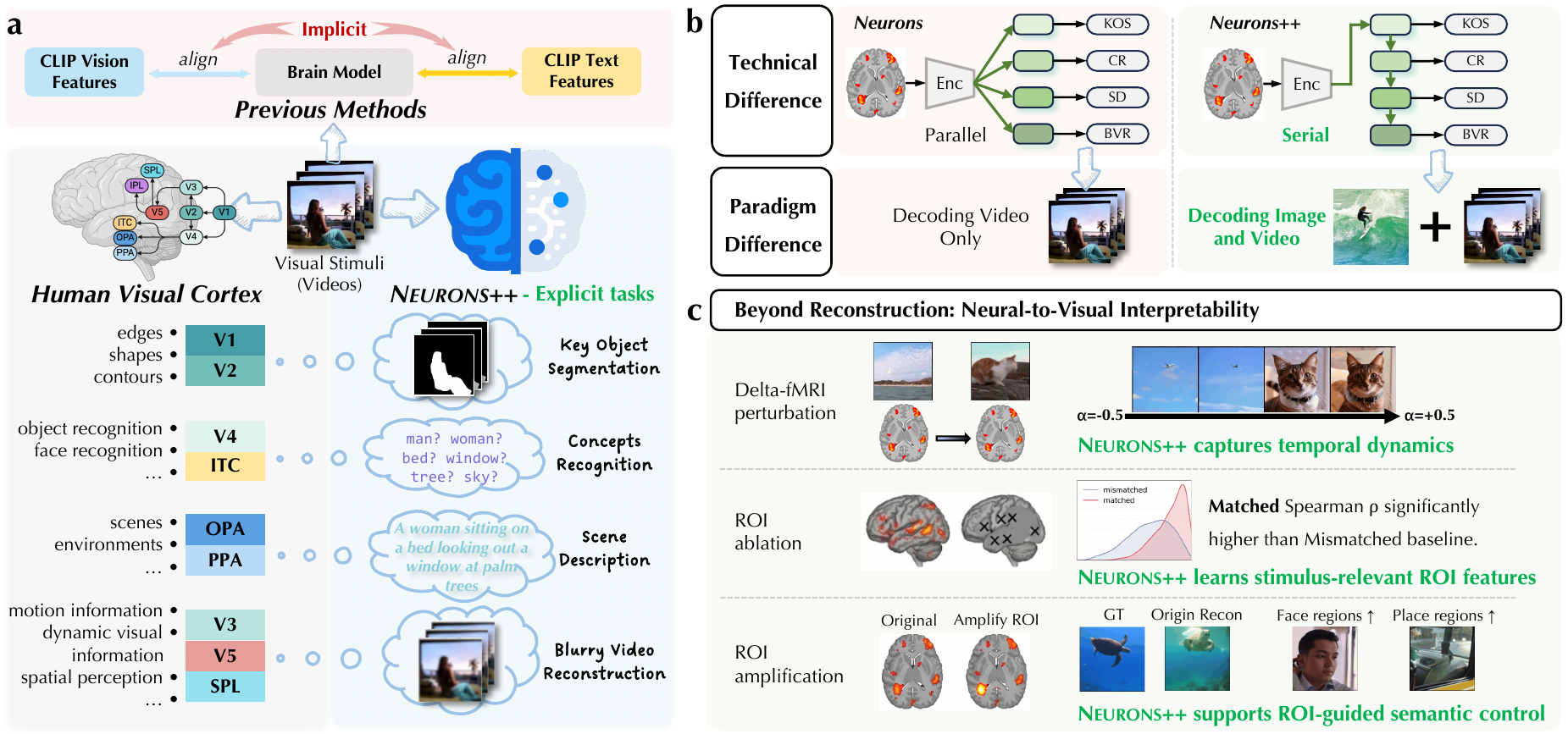}
\caption{
Overview and motivation of \methodname{}.
(a) Existing approaches often rely on implicit latent-space alignment or independently aggregated visual cues, leaving the neural-to-visual hierarchy weakly structured.
(b) \methodname{} connects static and dynamic reconstruction through a visual-stream formulation and progressively transforms neural activity into structured visual cues through a brain-inspired Serial Visual Decoupler.
(c) Delta-fMRI perturbation, ROI ablation, and ROI amplification provide evidence of directional controllability, stimulus-relevant cortical sensitivity, and region-selective semantic modulation.
}
\label{fig:intro}
\end{figure*}

To overcome these limitations, unlike previous methods that treat fMRI-to-image and fMRI-to-video reconstruction as separate tasks, we introduce a novel fMRI-to-visual-stream reconstruction setting and develop a novel framework, \methodname{}, to realize it. \textbf{To bridge the separated reconstruction settings}, we formulate images as single-frame visual streams and videos as multi-frame visual streams under a common neural-to-visual formulation, enabling multi-subject image-fMRI pretraining to provide stable spatial-semantic grounding and video-fMRI adaptation to extend the learned representation to temporally evolving visual experience. \textbf{To build a connected representation hierarchy}, we develop a novel Serial Visual Decoupler that progressively transforms neural activity into spatial object structure, object-level semantics, scene-level language, and coarse temporal appearance, forming a brain-inspired and traceable neural-to-cue pathway rather than independent generation conditions. \textbf{To verify neural grounding beyond visual fidelity}, we conduct delta-fMRI perturbation, ROI ablation, and ROI amplification experiments, showing that the decoded content changes directionally under graded neural perturbations, is sensitive to stimulus-relevant cortical features, and can be modulated through region-specific neural interventions.

We evaluate \methodname{} on fMRI-to-image and fMRI-to-video benchmarks, where it achieves state-of-the-art or competitive performance across low-level, semantic-level, and spatiotemporal metrics.

Compared with our conference version \textsc{Neurons}~\cite{wang2025neurons}, this journal extension makes three main scientific contributions:
\begin{itemize}
    \item We establish a new visual-stream formulation that connects static image reconstruction and dynamic video reconstruction under a common neural-to-visual setting.
    \item We introduce a novel brain-inspired Serial Visual Decoupler that organizes spatial, semantic, language-level, and temporal cues into a progressive neural-to-visual hierarchy.
    \item We provide experimental evidence that the decoded content is directionally controllable, sensitive to stimulus-relevant cortical activity, and modulated by region-specific neural interventions.
\end{itemize}
Together, these advances move fMRI reconstruction beyond task-specific visual generation toward structured and verifiable neural visual decoding.

\section{Related Work}

\subsection{fMRI-to-Image Reconstruction}

Reconstructing visual stimuli from fMRI has been widely studied for understanding neural visual representations and developing brain-computer interface systems. Early studies mainly focused on decoding visual categories or reconstructing simple visual patterns from brain activity~\cite{horikawa2017firstfmri}. With the emergence of large-scale fMRI-image datasets, especially the Natural Scenes Dataset (NSD)~\cite{NSD}, fMRI-to-image reconstruction has rapidly advanced from low-level image recovery to natural image generation.
Recent diffusion-based methods substantially improve reconstruction quality by aligning fMRI representations with vision-language models and using generative priors for image synthesis~\cite{takagi2023high,scotti2023reconstructing,mindeye2,gong2025mindtuner,fang2023alleviating,zengControllableMindVisual2024,chen2023seeing,huo2024neuropictor,ferrante2023brain,BRAIN_lu2025cognitive,BRAIN_wang2026zebra,brainit}. These methods typically map fMRI signals into CLIP~\cite{radford2021clip}, Stable Diffusion~\cite{rombach2022ldm}, or related latent spaces. To preserve low-level spatial details, several methods introduce additional control signals such as blurry reconstructions, contours, or masks~\cite{mindeye2,gong2025mindtuner,zengControllableMindVisual2024,ferrante2023brain}. To improve high-level semantic alignment, other methods leverage captions, category labels, or text embeddings~\cite{mindeye2,gong2025mindtuner,scotti2023reconstructing}. Cross-subject reconstruction methods further attempt to reduce subject-specific data requirements by aligning fMRI representations across individuals through shared latent spaces or functional alignment~\cite{wang2024mindbridge,thualAligningBrainFunctions2023,xia2024umbrae,qian2023fmri,li2024enhancing,scotti2023reconstructing}.
Despite these advances, existing fMRI-to-image methods are designed for static visual stimuli. They are usually optimized for spatial fidelity, object identity, and image-level semantics, but do not explicitly model temporal evolution, motion, or scene transitions. As a result, they cannot directly handle dynamic visual experiences such as videos.

\subsection{fMRI-to-Video Reconstruction}

Compared with static image reconstruction, fMRI-to-video reconstruction is more challenging because it requires recovering object motion, temporal consistency, and scene transitions from slow BOLD signals. Early studies explored dynamic visual decoding with handcrafted features or low-resolution reconstruction strategies~\cite{wen2018neural,wang2022reconstructing,kupershmidt2022penny}. Recent methods improve video reconstruction by aligning fMRI representations with vision-language embeddings and using generative video priors~\cite{chen2024cinematic,gong2025neuroclips,li2024enhancing}. These methods typically follow a two-stage paradigm: mapping fMRI signals into CLIP or related latent spaces, and then synthesizing videos with diffusion-based generative models~\cite{rombach2022ldm,guo2023animatediff,blattmann2023alignyourlatent}.
Recent state-of-the-art methods further introduce more structured guidance for video reconstruction. SemVideo~\cite{yang2026semvideo} uses hierarchical semantic guidance, including static anchors, motion narratives, and holistic summaries, to improve object consistency and temporal coherence. \textsc{Neurons}~\cite{wang2025neurons} introduces cortex-inspired explicit tasks to decompose fMRI-to-video reconstruction into multiple visual cues. However, these methods remain video-specific and do not connect static image reconstruction with dynamic video reconstruction under a shared visual-stream formulation. In addition, the explicit cues in \textsc{Neurons} are mainly aggregated in parallel, which limits their ability to form a progressive neural-to-visual pathway.

\subsection{Neural-to-Visual Interpretability}

Most generative brain decoding methods are primarily evaluated by reconstruction quality, including pixel-level similarity, semantic retrieval, feature similarity, or temporal consistency. These metrics are useful for measuring output fidelity, but they do not directly verify whether the generated content is faithfully grounded in brain activity. This issue becomes more important as powerful diffusion priors can hallucinate plausible visual details even when neural evidence is weak or incomplete.
Several existing methods improve controllability or introduce intermediate visual cues, such as masks, captions, concepts, or blurry reconstructions~\cite{mindeye2,gong2025mindtuner,zengControllableMindVisual2024,wang2025neurons,yang2026semvideo}. However, these cues are usually used as auxiliary conditions for generation, rather than as objects of neural-to-visual verification. In contrast, \methodname{} formulates image and video reconstruction as unified fMRI-to-visual-stream decoding, uses serial cortex-inspired decoding to progressively recover spatial, semantic, language-level, and temporal cues, and further examines brain-grounding through delta-fMRI perturbation, ROI ablation, and ROI amplification.
\section{Method}

\subsection{fMRI-to-Visual-Stream Problem Formulation}
\label{sec:problem}

We formulate fMRI-to-image and fMRI-to-video reconstruction as an fMRI-to-visual-stream decoding problem.
Given an fMRI response $\mathbf{x}_c$ corresponding to a perceived visual stimulus, the target visual stream is denoted as $\mathbf{Y}_c = \{ I_{c,t} \}_{t=1}^{T_c}$, where $I_{c,t}$ is the $t$-th visual frame and $T_c$ is the temporal length of the visual stream.
When $T_c=1$, the task corresponds to static image reconstruction; when $T_c>1$, the task corresponds to video reconstruction.
This formulation unifies image and video decoding at the level of visual-stream representation, without assuming that all image and video datasets are jointly trained by a single decoder.
Here, ``unified'' refers to a common visual-stream formulation, shared architectural design, and consistent inference interface for both image and video reconstruction.
Because fMRI responses are highly subject- and dataset-dependent, we perform subject-specific adaptation for each target dataset while keeping the same decoding formulation and module design.

For video datasets, each video is split into short clips according to the temporal resolution of fMRI.
In our implementation, each clip contains $F$ frames, i.e., $T_c=F$, and a single visual stream is represented as $\mathbf{Y}_c \in \mathbb{R}^{F\times C\times H\times W}$, where $F=6$ unless otherwise specified.
For image datasets, we set $T_c=1$, so the same formulation naturally reduces to single-frame reconstruction.
The goal of \methodname{} is to learn fMRI-derived representations that can be instantiated for both single-frame image generation and multi-frame video generation.

\begin{figure*}[t]
    \centering
    \includegraphics[width=1\linewidth]{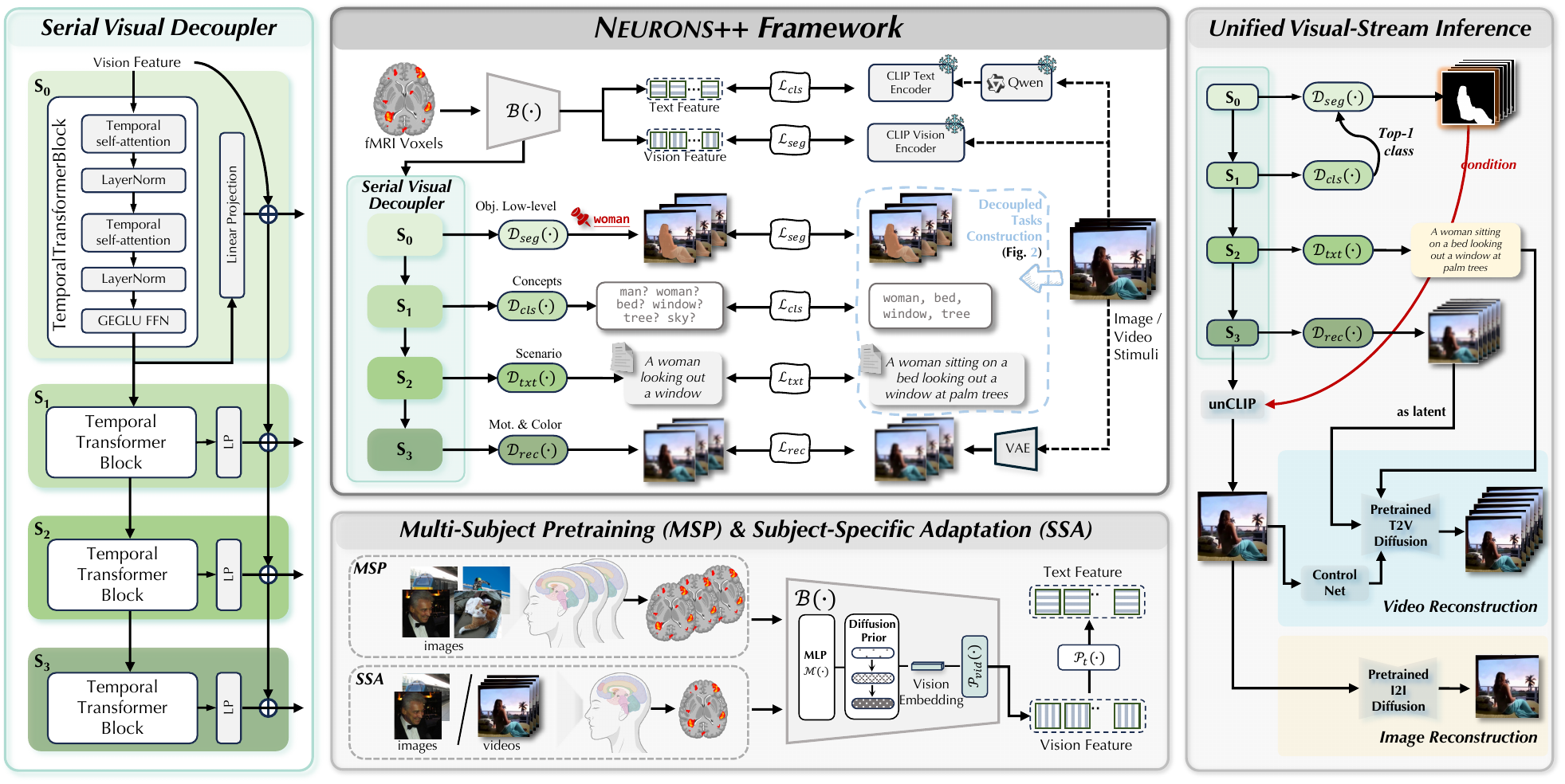}
    \caption{
    Overall framework of \methodname{}.
    The brain encoding module maps fMRI voxels into visual and textual embeddings aligned with foundation-model spaces.
    \methodname{} is trained through multi-subject pretraining (MSP) on image-fMRI data and subject-specific adaptation (SSA) on target image or video datasets.
    The Serial Visual Decoupler progressively decodes explicit visual cues through four stages: key object segmentation, concept recognition, scene description, and blurry visual-stream reconstruction.
    During inference, these decoded cues condition an unCLIP-based keyframe reconstruction process, followed by either a pretrained image-to-image diffusion model for image reconstruction or a pretrained text-to-video diffusion model for video reconstruction.
    }
    \label{fig:framework}
\end{figure*}

\begin{figure}[t]
    \centering
    \includegraphics[width=1\linewidth]{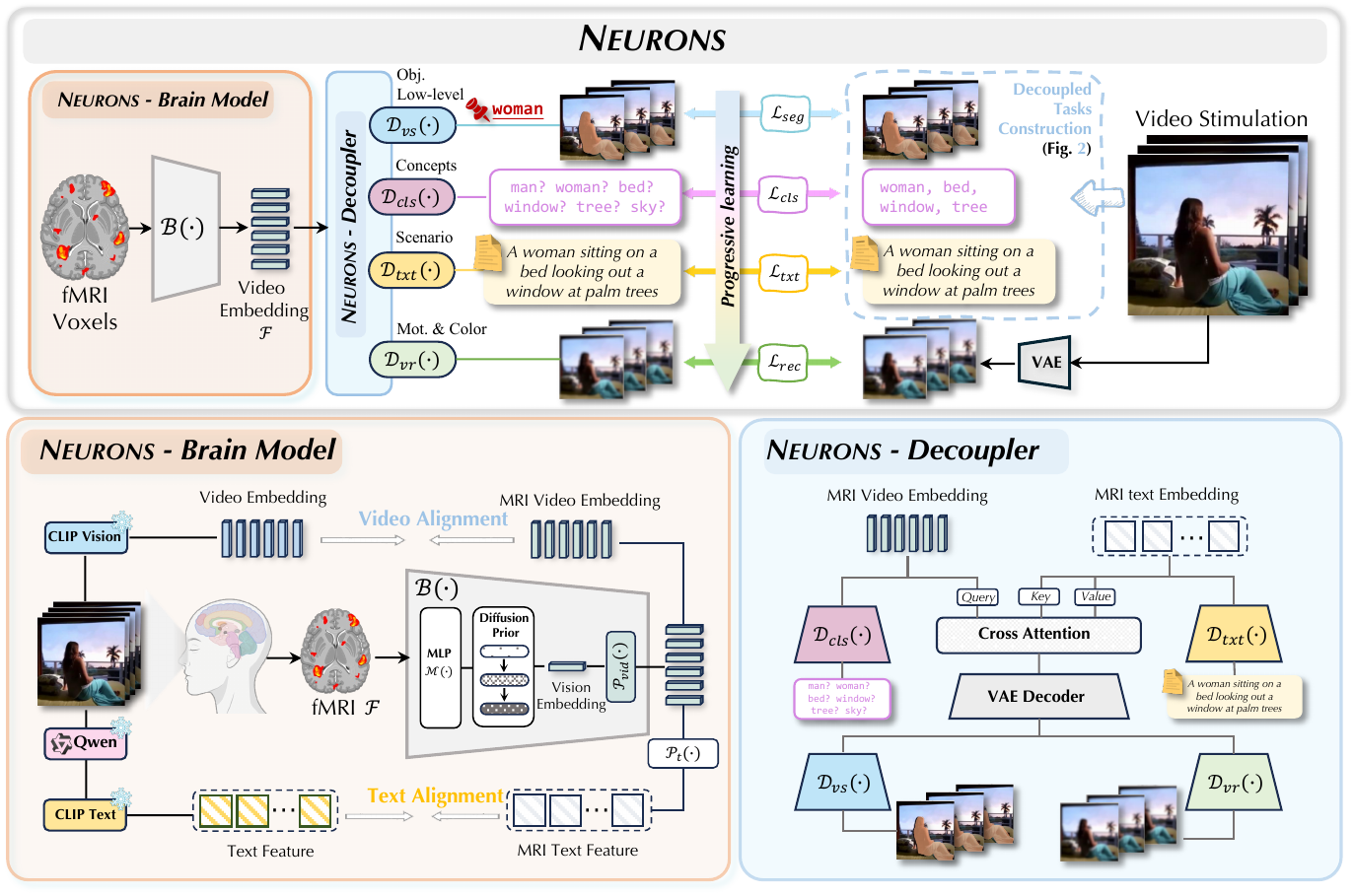}
    \caption{
    Overview of the conference version \confname{}~\cite{wang2025neurons}.
    In this journal version, we inherit the decoupled task construction while extending the setting from video-only reconstruction to visual-stream reconstruction and the architecture from parallel task aggregation to serial visual decoding.
    }
    \label{fig:framework_conf}
\end{figure}

\noindent\textbf{Notation.}
Throughout the paper, the subscript $c$ denotes an individual visual stream, i.e., a single fMRI--stimulus sample; the sample indices $i,j$ used in the interpretability analyses follow the same convention, i.e., $\mathbf{x}_i\equiv\mathbf{x}_{c=i}$.
We present the forward pass for a single stream, and \emph{all forward-pass tensors are written without a batch axis}: $\mathbf{e}^{i}_c\in\mathbb{R}^{N_{\mathrm{tok}}\times D}$, $\mathbf{e}^{v}_c\in\mathbb{R}^{T_c\times N_{\mathrm{tok}}\times D}$, $\mathbf{z}^{k}_c\in\mathbb{R}^{T_c\times N_{\mathrm{tok}}\times D}$, and $\hat{\mathbf{R}}_c\in\mathbb{R}^{T_c\times C\times H'\times W'}$ all refer to one stream.
The mini-batch index is introduced only where it is actually needed---in the contrastive objective and in the batch-averaged losses---as a superscript $(b)$ with $b\in\{1,\dots,B\}$ and $\cdot^{(b)}\equiv\cdot_{c=b}$ (e.g., $\hat{\mathbf{R}}^{(b)}$ is the prediction for the $b$-th stream of the batch).
We write $C$ (italic) for the latent channel dimension, indexed by $\gamma$ to avoid collision with the sample subscript $c$, and reserve the boldface symbol $\mathbf{C}_c$ for the concept-label vector of the recognition target in Eq.~(\ref{eq:cls}); these are distinct quantities.
Finally, $N_{\mathrm{tok}}$ denotes the number of CLIP visual tokens, which is separate from the candidate-class count $N$ of the $N$-way recognition metrics defined in Supplementary~\S\ref{sec:supp_metrics}.

\subsection{Overview of \methodname{}}
\label{sec:overview}

Our conference model, \confname{}~\cite{wang2025neurons} (Fig.~\ref{fig:framework_conf}), introduced a cortex-inspired fMRI-to-video reconstruction framework that decomposes decoding into four explicit sub-tasks: key object segmentation, concept recognition, scene description, and blurry video reconstruction.
Building on the visual-stream formulation introduced in \S\ref{sec:problem}, \methodname{} extends \confname{} from video-only reconstruction to both single-frame image reconstruction and multi-frame video reconstruction under a shared architectural template.
Image and video datasets are adapted separately at the subject level, while multi-subject image-fMRI pretraining provides a shared spatial-semantic initialization for both settings.
As shown in Fig.~\ref{fig:framework}, \methodname{} consists of three components.
First, a brain encoding module maps voxel responses into visual and textual embeddings aligned with vision-language foundation models (\S\ref{sec:brain_model}).
Second, a Serial Visual Decoupler converts the inherited tasks into a progressive decoding process, producing spatial, semantic, language-level, and temporal cues (\S\ref{sec:decoupler}).
Third, a unified visual-stream inference pipeline uses these cues for either single-frame image reconstruction or multi-frame video reconstruction (\S\ref{sec:infer}).
The construction of decoupled task labels follows \confname{} and is described in Supplementary~\S\ref{sec:supp_dtc}.

\subsection{Multi-Subject Pretraining \& Subject-Specific Adaptation}
\label{sec:brain_model}

\methodname{} is trained in two stages to transfer spatial-semantic representations from image-fMRI data to static and dynamic visual-stream reconstruction.
In the first stage, the trainable model components are pretrained on image-fMRI data across eight subjects.
This multi-subject pretraining provides a stable spatial-semantic initialization because static image stimuli offer cleaner supervision for object identity, spatial layout, and scene semantics.
In the second stage, the pretrained weights are loaded for subject-specific adaptation on each target image or video dataset.

The brain encoding module maps voxel responses into visual and textual embeddings for visual-stream reconstruction.
Given a preprocessed fMRI response $\mathbf{x}_c\in\mathbb{R}^{V}$, the brain encoder applies a ridge regression module $\mathrm{ridge}(\cdot)$ that reduces the voxel dimension, followed by a residual-MLP backbone $\mathcal{B}_{\theta}(\cdot)$ (the MindEye2-style encoder~\cite{mindeye2}), yielding the neural representation $\mathbf{h}_c = \mathcal{B}_{\theta}(\mathrm{ridge}(\mathbf{x}_c))$.
Throughout, $\mathcal{B}_{\theta}$ denotes the residual-MLP backbone only, excluding the ridge stage.
The detailed architecture is provided in Supplementary~\S\ref{sec:supp_BMdetail}.

To align the fMRI-derived representation with foundation-model spaces, we project $\mathbf{h}_c$ into both visual and textual embeddings.
An image-level projection head $\mathcal{P}_{i}(\cdot)$ maps $\mathbf{h}_c$ into visual tokens, $\mathbf{e}^{i}_c = \mathcal{P}_{i}(\mathbf{h}_c)$, where $\mathbf{e}^{i}_c \in \mathbb{R}^{N_{\mathrm{tok}}\times D}$, $N_{\mathrm{tok}}$ denotes the number of visual tokens, and $D$ is the feature dimension.
This image-level head, the backbone, and the ridge map compose into the read-out $f=\mathcal{P}_{i}\circ\mathcal{B}_{\theta}\circ\mathrm{ridge}$ that we probe in the interpretability analysis (\S\ref{subsec:roi_ablation}).
A temporal projection module $\mathcal{P}_{vid}(\cdot)$ then expands the image-level representation into a visual-stream representation, $\mathbf{e}^{v}_c = \mathcal{P}_{vid}(\mathbf{e}^{i}_c; T_c)$, where $\mathbf{e}^{v}_c \in \mathbb{R}^{T_c\times N_{\mathrm{tok}}\times D}$.
For image reconstruction, $T_c=1$; for video reconstruction, $T_c>1$.
In parallel, a text projection head produces an fMRI-derived text embedding $\mathbf{e}^{t}_c = \mathcal{P}_{t}(\mathbf{h}_c)$.
Following the MindEye2-style design~\cite{mindeye2}, a diffusion prior $\mathcal{M}(\cdot)$ additionally maps the pooled fMRI-derived visual embedding into the CLIP visual-embedding space used by the downstream generative prior, and is trained with a diffusion-prior loss $\mathcal{L}_{prior}$.
Here $\mathcal{M}(\cdot)$ denotes the diffusion-prior \emph{module}, and its internal noise-prediction network---the network actually optimized by $\mathcal{L}_{prior}$ in Eq.~(\ref{eq:enc})---is denoted $\epsilon_{\psi}(\cdot)$. We use $\mathcal{M}(\cdot)$ when referring to the prior as a module (Fig.~\ref{fig:framework}) and $\epsilon_{\psi}(\cdot)$ for its trained noise predictor.

Let $\phi_v(I_{c,t}) \in \mathbb{R}^{N_{\mathrm{tok}}\times D}$ denote the CLIP visual-token representation of frame $I_{c,t}$.
For a visual stream $\mathbf{Y}_c$, we define the frame-wise target visual representation as
\begin{equation} 
    \Phi_v(\mathbf{Y}_c)=\{\phi_v(I_{c,t})\}_{t=1}^{T_c}\in\mathbb{R}^{T_c\times N_{\mathrm{tok}}\times D}.
\end{equation}
During contrastive alignment, both the predicted visual-stream embedding $\mathbf{e}^{v}_c$ and the target representation $\Phi_v(\mathbf{Y}_c)$ are mean-pooled over the temporal and token dimensions before applying the BiMixCo objective.
Concretely, for a single stream $\mathrm{Pool}(\cdot):\mathbb{R}^{T_c\times N_{\mathrm{tok}}\times D}\to\mathbb{R}^{D}$ averages over the $T_c$ temporal positions and the $N_{\mathrm{tok}}$ tokens, leaving only the feature axis. For a single visual stream,
\begin{equation}
\label{eq:pool}
\begin{aligned}
    \bar{\mathbf{e}}^{v}_c
    &=
    \frac{1}{T_c N_{\mathrm{tok}}}
    \sum_{t=1}^{T_c}\sum_{n=1}^{N_{\mathrm{tok}}}
    \mathbf{e}^{v}_{c,t,n}
    \in\mathbb{R}^{D},\\
    \bar{\Phi}_v(\mathbf{Y}_c)
    &=
    \frac{1}{T_c N_{\mathrm{tok}}}
    \sum_{t=1}^{T_c}\sum_{n=1}^{N_{\mathrm{tok}}}
    \phi_v(I_{c,t})_{n}
    \in\mathbb{R}^{D}.
\end{aligned}
\end{equation}
where $\mathbf{e}^{v}_{c,t,n}\in\mathbb{R}^{D}$ is the feature vector at temporal position $t$ and token $n$, and $\phi_v(I_{c,t})_{n}\in\mathbb{R}^{D}$ is the $n$-th CLIP token of frame $I_{c,t}$.
Over a mini-batch of $B$ visual streams, the pooled embeddings stack into $\mathbb{R}^{B\times D}$ before the contrastive objective is applied.
Let $\Phi_t(\mathbf{S}_c)$ denote the target text embedding extracted by the CLIP text encoder, where $\mathbf{S}_c$ is the scene description generated during decoupled task construction.
The embedding-alignment objective combines the visual and textual CLIP alignment losses with the diffusion-prior loss:
\begin{equation}
\label{eq:enc}
\begin{aligned}
    \mathcal{L}_{\mathrm{enc}}
    ={}&
    \mathcal{L}_{\mathrm{CLIP}_v}
    \big(\bar{\mathbf{e}}^{v}_c, \bar{\Phi}_v(\mathbf{Y}_c)\big)\\
    &+
    \lambda_t
    \mathcal{L}_{\mathrm{CLIP}_t}
    \big(\mathbf{e}^{t}_c, \Phi_t(\mathbf{S}_c)\big)\\
    &+
    \lambda_p
    \mathcal{L}_{prior},
\end{aligned}
\end{equation}
where $\lambda_t$ and $\lambda_p$ balance the text-alignment and diffusion-prior terms.
Following~\cite{mindeye2,gong2025neuroclips}, $\mathcal{L}_{\mathrm{CLIP}_v}$ and $\mathcal{L}_{\mathrm{CLIP}_t}$ are implemented with the BiMixCo contrastive objective, which combines MixCo augmentation and bidirectional contrastive learning.
The diffusion-prior loss $\mathcal{L}_{prior}$ trains the prior network $\epsilon_{\psi}(\cdot)$ to denoise the target CLIP visual embedding conditioned on the fMRI-derived embedding, i.e., $\mathcal{L}_{prior}=\mathbb{E}_{c,\,s,\,\boldsymbol{\epsilon}}\big[\lVert\boldsymbol{\epsilon}-\epsilon_{\psi}(\tilde{\mathbf{e}}^{s}_{y,c},s,\mathbf{e}_{x,c})\rVert_2^2\big]$, where $\mathbf{e}_{x,c}=\bar{\mathbf{e}}^{v}_c$ is the pooled fMRI-derived visual embedding, $\mathbf{e}_{y,c}=\bar{\Phi}_v(\mathbf{Y}_c)$ is the pooled CLIP visual target, $\tilde{\mathbf{e}}^{s}_{y,c}=\sqrt{\bar{\alpha}_s}\,\mathbf{e}_{y,c}+\sqrt{1-\bar{\alpha}_s}\,\boldsymbol{\epsilon}$ is the step-$s$ noised target, and $\boldsymbol{\epsilon}\sim\mathcal{N}(\mathbf{0},\mathbf{I})$ is the injected noise.
The cumulative noise schedule $\bar{\alpha}_s$ follows the standard DDPM definition and is specified in Supplementary~\S\ref{sec:supp_BMdetail}.
The resulting visual and textual embeddings are then fed into the Serial Visual Decoupler.

\subsection{Serial Visual Decoupler}
\label{sec:decoupler}

The Serial Visual Decoupler transforms the fMRI-derived visual-stream
embedding into explicit visual cues through a progressive decoding process.
Unlike the conference version, where the four task heads operate in parallel,
\methodname{} organizes them serially: low-level object structure is decoded
first, followed by object-level concepts, scene-level language description,
and finally coarse visual-stream reconstruction. This design encourages later
stages to inherit previously decoded information while retaining access to
the original fMRI-derived visual representation.

Given the visual-stream embedding $\mathbf{e}^{v}_c$, we initialize
$\mathbf{z}^{-1}_c=\mathbf{e}^{v}_c$ and apply four temporal-transformer
stages S$_0$--S$_3$. Each stage updates the representation as
\begin{equation}
\label{eq:residual}
    \mathbf{z}^{k}_c
    =
    \mathcal{V}_{k}(\mathbf{z}^{k-1}_c, \mathbf{e}^{v}_c)
    =
    \mathbf{z}^{k-1}_c
    +
    \mathrm{TTB}_{k}(\mathbf{z}^{k-1}_c)
    +
    \mathrm{LP}_{k}(\mathbf{e}^{v}_c),
\end{equation}
where $\mathrm{TTB}_{k}(\cdot)$ is a temporal transformer block and
$\mathrm{LP}_{k}(\cdot)$ is a linear projection. All stage representations
preserve the shape of the input visual-stream embedding,
$\mathbf{z}^{k}_c\in\mathbb{R}^{T_c\times N_{\mathrm{tok}}\times D}$.

The four stages are associated with four explicit visual cues:
\begin{equation}
\begin{aligned}
    \mathbf{z}^{0}_c &\rightarrow \text{key object segmentation}, \\
    \mathbf{z}^{1}_c &\rightarrow \text{concept recognition}, \\
    \mathbf{z}^{2}_c &\rightarrow \text{scene description}, \\
    \mathbf{z}^{3}_c &\rightarrow \text{blurry visual-stream reconstruction}.
\end{aligned}
\end{equation}
These supervision targets are inherited from \confname{} and are constructed
as described in Supplementary~\S\ref{sec:supp_dtc}. The task-head
architectures and loss definitions are provided in
Supplementary~\S\ref{sec:supp_decoupler_losses}.
The decoupling objective is
\begin{equation}
    \mathcal{L}_{dec}
    =
    \mathcal{L}_{seg}
    +
    \mathcal{L}_{cls}
    +
    \mathcal{L}_{txt}
    +
    \mathcal{L}_{rec}.
\end{equation}
During multi-subject pretraining and subject-specific adaptation, the full
trainable model is optimized with
\begin{equation}
\label{eq:total}
    \mathcal{L}
    =
    \mathcal{L}_{enc}
    +
    \lambda_{dec}\mathcal{L}_{dec},
\end{equation}
where $\mathcal{L}_{enc}$ is the embedding-alignment objective and
$\lambda_{dec}$ balances neural embedding alignment with explicit visual-cue
decoding. Implementation details and loss weights are reported in
Supplementary~\S\ref{sec:supp_training_config}.

\subsection{Unified Visual-Stream Inference}
\label{sec:infer}

During inference, \methodname{} reconstructs visual streams through the serial decoding process.
Given a test fMRI response $\mathbf{x}_c$, the brain encoding module first produces visual and textual embeddings, which are then passed to the Serial Visual Decoupler.
The decoupler progressively predicts four explicit cues: a key object mask, a concept set, a scene description, and a blurry visual-stream reconstruction.
These cues provide spatial, semantic, language-level, and temporal guidance for final generative reconstruction.
We first reconstruct a keyframe using an unCLIP-based image generation pipeline~\cite{ramesh2022hierarchical}.
The keyframe is conditioned on the fMRI-derived visual embedding, the decoded scene description, and the predicted key object mask.
The binary key object mask is rescaled to $[0.5,1]$ and used as a soft spatial condition, emphasizing the salient object while preserving background context.
For image reconstruction, the keyframe is further refined by a pretrained SDXL-based image-to-image diffusion model, producing the final single-frame visual-stream reconstruction; the specific checkpoint and img2img settings are reported in Supplementary~\S\ref{sec:supp_training_config}.
For video reconstruction, the keyframe is expanded into a multi-frame visual stream using a pretrained text-to-video diffusion model~\cite{guo2023animatediff}.
The video generator is conditioned on the reconstructed keyframe, decoded scene description, predicted concepts, object mask, and blurry visual-stream reconstruction.
In this way, image and video reconstruction share the same neural-to-cue decoding interface, while the final visual rendering is performed by modality-appropriate pretrained generative priors for single-frame and multi-frame visual-stream generation.

\begin{table*}[ht]
\centering
\caption{
Unified quantitative comparison across fMRI-to-video and fMRI-to-image reconstruction benchmarks.
CC2017 and HCP report video reconstruction metrics, while NSD reports image reconstruction metrics.
Video metrics are reported as percentages.
Results on CC2017 and HCP are averaged over three subjects; NSD results are averaged over subjects 1, 2, 5, and 7.
`$\pm$' denotes per-sample standard deviation across pooled test clips for video datasets.
\redx{} indicates that the method does not support the corresponding modality, while \NR{} indicates that the metric is not reported for an otherwise compatible modality.
For NSD image reconstruction, we report a compact subset of representative low- and high-level metrics in the main table; the full set of NSD metrics is reported in Supplementary~Table~\ref{tab:image_nsd_full}.
Bold and underlined values indicate the best and second-best performance within each metric, respectively.
Asterisks mark metrics on which \methodname{} significantly improves over the comparison baseline:
SemVideo on CC2017 and \textsc{Neurons} on HCP
($^{\ast}p{<}0.05$, $^{\ast\ast}p{<}0.01$, $^{\ast\ast\ast}p{<}0.001$).
}
\begingroup
\setlength{\tabcolsep}{2.0pt}
\renewcommand{\arraystretch}{1.03}
\resizebox{\textwidth}{!}{
\begin{tabular}{@{}lccc|ccc|cccc@{}}
\toprule
\multirow{3}{*}{Method}
& \multicolumn{3}{c|}{fMRI-to-Video (CC2017)}
& \multicolumn{3}{c|}{fMRI-to-Video (HCP)}
& \multicolumn{4}{c}{fMRI-to-Image (NSD)} \\
\cmidrule(lr){2-4}
\cmidrule(lr){5-7}
\cmidrule(lr){8-11}
& \multicolumn{2}{c}{Semantic-level} & ST-level
& \multicolumn{2}{c}{Semantic-level} & ST-level
& \multicolumn{2}{c}{Low-Level}
& \multicolumn{2}{c}{High-Level} \\
\cmidrule(lr){2-3}
\cmidrule(lr){4-4}
\cmidrule(lr){5-6}
\cmidrule(lr){7-7}
\cmidrule(lr){8-9}
\cmidrule(lr){10-11}
& 2-way$\uparrow$ & 50-way$\uparrow$ & CLIP-pcc$\uparrow$
& 2-way$\uparrow$ & 50-way$\uparrow$ & CLIP-pcc$\uparrow$
& PixCorr$\uparrow$ & Alex(5)$\uparrow$
& Incep$\uparrow$ & Eff$\downarrow$ \\
\midrule

Wen~\cite{wen2018neural} \scriptsize\textcolor{gray}{[Cereb. Cortex'18]}
& \NR & 16.6\%{\scriptsize$\pm2\%$} & \NR
& \NR & \NR & \NR
& \redx & \redx & \redx & \redx \\

Wang~\cite{wang2022reconstructing} \scriptsize\textcolor{gray}{[Cereb. Cortex'22]}
& 77.3\%{\scriptsize$\pm3\%$} & \NR & 40.2\%{\scriptsize$\pm41\%$}
& \NR & \NR & \NR
& \redx & \redx & \redx & \redx \\

MinD-Video~\cite{chen2024cinematic} \scriptsize\textcolor{gray}{[NeurIPS'24]}
& 83.9\%{\scriptsize$\pm3\%$} & 19.7\%{\scriptsize$\pm2\%$} & 40.8\%{\scriptsize$\pm46\%$}
& 76.9\%{\scriptsize$\pm2\%$} & \NR & 49.9\%{\scriptsize$\pm3\%$}
& \redx & \redx & \redx & \redx \\

NeuroClips~\cite{gong2025neuroclips} \scriptsize\textcolor{gray}{[NeurIPS'25]}
& 83.4\%{\scriptsize$\pm3\%$} & 22.0\%{\scriptsize$\pm1\%$} & 73.8\%{\scriptsize$\pm17\%$}
& \NR & \NR & \NR
& \redx & \redx & \redx & \redx \\

CineNeuron~\cite{cineneuron} \scriptsize\textcolor{gray}{[CVPR'26]}
& 85.0\%{\scriptsize$\pm2\%$} & 24.0\%{\scriptsize$\pm2\%$} & \NR
& \NR & \NR & \NR
& \redx & \redx & \redx & \redx \\

MindCross~\cite{liu2026mindcross} \scriptsize\textcolor{gray}{[AAAI'26]}
& 83.0\%{\scriptsize$\pm2\%$} & 21.0\%{\scriptsize$\pm2\%$} & 76.2\%{\scriptsize$\pm2\%$}
& \NR & \NR & \NR
& \redx & \redx & \redx & \redx \\

MindAnimate~\cite{lu2024animate} \scriptsize\textcolor{gray}{[ICLR'25]}
& \NR & \NR & \NR
& 77.8\%{\scriptsize$\pm2\%$} & \NR & 51.1\%{\scriptsize$\pm5\%$}
& \redx & \redx & \redx & \redx \\

\textsc{Neurons}~\cite{wang2025neurons} \scriptsize\textcolor{gray}{[ICCV'25]}
& 86.3\%{\scriptsize$\pm3\%$} & 26.2\%{\scriptsize$\pm1\%$} & 93.4\%{\scriptsize$\pm17\%$}
& \underline{82.1\%{\scriptsize$\pm3\%$}} & \underline{16.7\%{\scriptsize$\pm3\%$}} & \underline{91.9\%{\scriptsize$\pm8\%$}}
& \redx & \redx & \redx & \redx \\

SemVideo~\cite{yang2026semvideo} \scriptsize\textcolor{gray}{[CVPR'26]}
& \underline{86.5\%{\scriptsize$\pm2\%$}} & \underline{26.4\%{\scriptsize$\pm2\%$}} & \NR
& 79.6\% & \NR & 52.3\%
& \redx & \redx & \redx & \redx \\

\midrule

Takagi...~\cite{takagi2023high} \scriptsize\textcolor{gray}{[CVPR'23]}
& \redx & \redx & \redx
& \redx & \redx & \redx
& 0.246 & 85.6\% & 83.8\% & 0.811 \\

MindEye1~\cite{scotti2023reconstructing} \scriptsize\textcolor{gray}{[NeurIPS'24]}
& \redx & \redx & \redx
& \redx & \redx & \redx
& 0.319 & 96.9\% & 94.6\% & 0.648 \\

UMBRAE~\cite{xia2024umbrae} \scriptsize\textcolor{gray}{[ECCV'24]}
& \redx & \redx & \redx
& \redx & \redx & \redx
& 0.283 & 97.0\% & 91.7\% & 0.700 \\

NeuroPictor~\cite{huo2024neuropictor} \scriptsize\textcolor{gray}{[ECCV'24]}
& \redx & \redx & \redx
& \redx & \redx & \redx
& 0.229 & 98.4\% & 94.5\% & 0.639 \\

MindBridge~\cite{wang2024mindbridge} \scriptsize\textcolor{gray}{[CVPR'24]}
& \redx & \redx & \redx
& \redx & \redx & \redx
& 0.151 & 95.5\% & 92.4\% & 0.712 \\

MindEye2~\cite{mindeye2} \scriptsize\textcolor{gray}{[ICML'24]}
& \redx & \redx & \redx
& \redx & \redx & \redx
& 0.322 & \underline{98.6\%} & 95.4\% & \underline{0.619} \\

MindTuner~\cite{gong2025mindtuner} \scriptsize\textcolor{gray}{[AAAI'25]}
& \redx & \redx & \redx
& \redx & \redx & \redx
& 0.322 & \textbf{98.8\%} & \underline{95.6\%} & \textbf{0.612} \\

BrainIT~\cite{brainit} \scriptsize\textcolor{gray}{[ICLR'26]}
& \redx & \redx & \redx
& \redx & \redx & \redx
& \underline{0.331} & \underline{98.6\%} & 94.4\% & 0.648 \\

\midrule

\rowcolor{cyan!10}
\textbf{\methodname{}}
& \textbf{87.1\%{\scriptsize$\pm3\%$}}$^{\ast\ast}$
& \textbf{27.9\%{\scriptsize$\pm1\%$}}$^{\ast}$
& \textbf{94.6\%{\scriptsize$\pm6\%$}}
& \textbf{85.0\%{\scriptsize$\pm3\%$}}$^{\ast\ast\ast}$
& \textbf{19.9\%{\scriptsize$\pm2\%$}}$^{\ast\ast\ast}$
& \textbf{93.2\%{\scriptsize$\pm6\%$}}$^{\ast\ast\ast}$
& \textbf{0.355}
& \textbf{98.8\%}
& \textbf{96.2\%}
& \textbf{0.612} \\

\bottomrule
\end{tabular}}
\endgroup
\label{tab:unified_comparison}
\end{table*}

\begin{figure}[t]
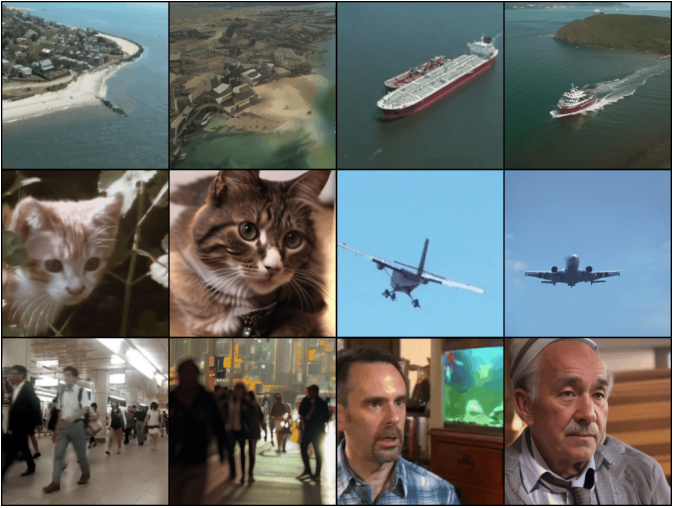

    \centering
    \animategraphics[width=0.5\linewidth]{8}{videos/cc2017_50/}{01}{06}%
    \animategraphics[width=0.5\linewidth]{8}{videos/cc2017_181/}{01}{06}\\

    \animategraphics[width=0.5\linewidth]{8}{videos/cc2017_295/}{01}{06}%
    \animategraphics[width=0.5\linewidth]{8}{videos/cc2017_395/}{01}{06}\\

    \animategraphics[width=0.5\linewidth]{8}{videos/cc2017_872/}{01}{06}%
    \animategraphics[width=0.5\linewidth]{8}{videos/cc2017_1195/}{01}{06}\\
    \caption{
    Qualitative comparison of fMRI-to-video reconstruction on the CC2017 dataset.
    For each example, the left clip is the ground-truth stimulus and the right clip is the reconstructed video; each row shows two examples.
    \textit{Best viewed with Acrobat Reader. Click the images to play the animation clips.}
    }
    \label{fig:cc2017_vis}
\end{figure}

\begin{figure}[t]
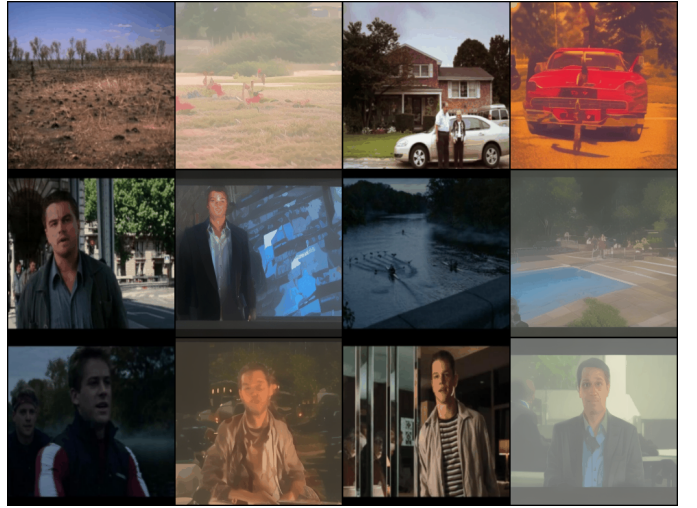

    \centering
    \animategraphics[width=0.5\linewidth]{8}{videos/hcp_1/020/}{01}{08}%
    \animategraphics[width=0.5\linewidth]{8}{videos/hcp_1/055/}{01}{08}\\

    \animategraphics[width=0.5\linewidth]{8}{videos/hcp_1/094/}{01}{08}%
    \animategraphics[width=0.5\linewidth]{8}{videos/hcp_1/115/}{01}{08}\\

    \animategraphics[width=0.5\linewidth]{8}{videos/hcp_1/118/}{01}{08}%
    \animategraphics[width=0.5\linewidth]{8}{videos/hcp_1/171/}{01}{08}\\
    \caption{Qualitative comparison of fMRI-to-video reconstruction on the HCP subject 1. For each example, the left clip is the ground-truth stimulus and the right clip is the reconstructed result; every row shows two examples. \textit{Best viewed with Acrobat Reader. Click the images to play the animation clips.}}
    \label{fig:hcp_vis}
\end{figure}

\section{Datasets and Implementation Details}

\subsection{Datasets}

\noindent\textbf{fMRI-to-Video Datasets.}
We evaluate fMRI-to-video reconstruction on two public video-fMRI datasets: CC2017~\cite{wen2018neural} and HCP~\cite{marcus2011informatics}. 
CC2017 contains fMRI responses from three subjects watching natural movie clips, with 18 training movies and 5 testing movies. 
Following prior work~\cite{wen2018neural,gong2025neuroclips}, we use the trial-averaged test responses and apply a hemodynamic lag when aligning BOLD signals with video stimuli.
For HCP, we use three subjects from the movie-watching task and follow the standard HCP preprocessing pipeline.
The visual cortical voxels are extracted using functionally defined ROIs. As there is no established train/test split for these subjects, we construct a leakage-free, group-aware temporal split that prevents temporally adjacent clips from straddling the train/test boundary (Supplementary~\S\ref{sec:supp_preprocessing}).
Detailed dataset statistics, voxel selection, BOLD-stimulus alignment, and preprocessing procedures are provided in Supplementary~\S\ref{sec:supp_preprocessing}.

\noindent\textbf{fMRI-to-Image Dataset.}
For fMRI-to-image reconstruction, we use the Natural Scenes Dataset (NSD)~\cite{NSD}, a large-scale public fMRI dataset containing brain responses to natural images from COCO~\cite{lin2014microsoft}. 
Following standard NSD reconstruction protocols~\cite{takagi2023high,mindeye2,gong2025mindtuner}, we evaluate on the shared test images and report results averaged over the subjects with complete scanning sessions. 
We use preprocessed fMRI responses from the visually responsive posterior cortex and adopt the same train/test split and normalization strategy as previous fMRI-to-image reconstruction methods. 
More details are provided in Supplementary~\S\ref{sec:supp_preprocessing}.

\begin{figure}[t]
    \centering
    \includegraphics[width=\linewidth]{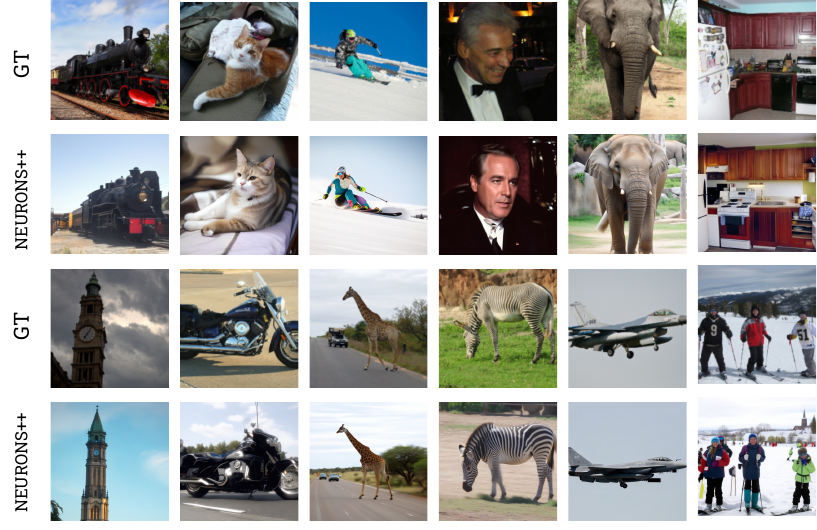}
    \caption{Qualitative comparison of fMRI-to-image reconstruction on NSD. Each pair shows the ground-truth image and the NEURONS++ reconstruction.}
    \label{fig:img_vis}
\end{figure}

\begin{table}[t]
  \centering
  \caption{
  Ablation study on the two main extensions from \confname{} to \methodname{} on video-based metrics.
  All results are reported on CC2017 Subject 1.
  `Base' denotes \confname{}, `MSP' denotes multi-subject image-fMRI pretraining, and `SVD' denotes the Serial Visual Decoupler.
  }
  \resizebox{0.8\linewidth}{!}{
  \begin{tabular}{ccc|ccc}
    \toprule
    \multicolumn{3}{c|}{Key Components}
    & \multicolumn{2}{c}{Semantic-level}
    & ST-level \\
    \cmidrule(lr){1-3}
    \cmidrule(lr){4-5}
    \cmidrule(lr){6-6}
    Base & MSP & SVD
    & 2-way$\uparrow$ & 50-way$\uparrow$ & CLIP-pcc$\uparrow$ \\
    \midrule

    \ding{51} &  & 
    & 86.2\% & 25.4\% & 93.2\% \\

    \ding{51} & \ding{51} & 
    & 86.5\% & 26.1\% & 93.8\% \\

    \ding{51} &  & \ding{51}
    & 86.3\% & 26.3\% & 96.7\% \\

    \midrule
    \rowcolor{cyan!10}
    \ding{51} & \ding{51} & \ding{51}
    & \textbf{86.5\%} & \textbf{26.6\%} & \textbf{97.2\%} \\

    \bottomrule
  \end{tabular}}
  \label{tab:ablation}
\end{table}

\begin{table*}[t]
  \centering
  \caption{
  Image-side ablation study on fMRI-to-image reconstruction on NSD Subject 1.
  `Base' denotes \confname{}, `MSP' denotes multi-subject image-fMRI pretraining, `SVD' denotes the Serial Visual Decoupler, and `I2I' denotes the image-to-image diffusion refinement stage.
  Bold and underlined values indicate the best and second-best performance, respectively.
  }
  \resizebox{0.7\textwidth}{!}{
  \begin{tabular}{cccc|cccc|cccc}
    \toprule
    \multicolumn{4}{c|}{Key Components}
    & \multicolumn{4}{c|}{Low-Level}
    & \multicolumn{4}{c}{High-Level} \\
    \cmidrule(lr){1-4}
    \cmidrule(lr){5-8}
    \cmidrule(lr){9-12}
    Base & MSP & SVD & I2I
    & PixCorr$\uparrow$
    & SSIM$\uparrow$
    & Alex(2)$\uparrow$
    & Alex(5)$\uparrow$
    & Incep$\uparrow$
    & CLIP$\uparrow$
    & Eff$\downarrow$
    & SwAV$\downarrow$ \\
    \midrule

    \ding{51} &  & \ding{51} & \ding{51}
    & 0.412 & 0.432 & 97.1\% & 98.3\% & 96.1\% & 93.6\% & 0.612 & 0.368 \\

    \ding{51} & \ding{51} &  & \ding{51}
    & 0.388 & \underline{0.463} & 95.8\% & 98.6\% & 95.9\% & 92.7\% & 0.628 & 0.374 \\

    \ding{51} & \ding{51} & \ding{51} & 
    & \underline{0.419} & 0.455 & \textbf{98.1\%} & \textbf{99.5\%} & \textbf{97.1\%} & \textbf{94.8\%} & \textbf{0.592} & \textbf{0.353} \\

    \midrule
    \rowcolor{cyan!10}
    \ding{51} & \ding{51} & \ding{51} & \ding{51}
    & \textbf{0.424} & \textbf{0.478} & \underline{98.0\%} & \underline{99.3\%}
    & \underline{96.7\%} & \underline{94.1\%} & \underline{0.606} & \underline{0.364} \\

    \bottomrule
  \end{tabular}}
  \label{tab:ablation_nsd}
\end{table*}

\subsection{Implementation Details and Evaluation Metrics}
We implement \methodname{} following standard preprocessing, training, and evaluation protocols from prior fMRI-to-image and fMRI-to-video reconstruction studies.
For videos, each stimulus is divided into short clips according to the fMRI temporal resolution, with BOLD responses aligned using a hemodynamic lag.
For images, each stimulus is treated as a single-frame visual stream under the same decoding formulation.
We evaluate videos with semantic-level retrieval/classification accuracy and spatiotemporal consistency metrics, and images with standard low- and high-level NSD reconstruction metrics.
Detailed preprocessing, model configurations, training hyperparameters, inference settings, and metric definitions are provided in Supplementary~\S\ref{sec:supp_implementation}\&\ref{sec:supp_metrics}.

\section{Experimental Evaluation}

\subsection{Comparison with State-of-the-Art Methods}

\noindent\textbf{Video Reconstruction.}
We first compare \methodname{} with representative fMRI-to-video reconstruction methods on CC2017 and HCP.
As shown in Table~\ref{tab:unified_comparison}, \methodname{} achieves the best overall performance on CC2017, improving over the conference version \confname{} from 0.863 to 0.871 in 2-way accuracy, from 0.262 to 0.279 in 50-way accuracy, and from 0.934 to 0.946 in CLIP-pcc.
Qualitative comparisons with ground-truth videos are shown in Fig.~\ref{fig:cc2017_vis}.
On HCP, which we use as a complementary cross-dataset check under a leakage-free, group-aware temporal split (Supplementary~\S\ref{sec:supp_preprocessing}), \methodname{} likewise improves semantic-level accuracy over the conference version \confname{} while maintaining comparable spatiotemporal consistency (Table~\ref{tab:unified_comparison}), confirming that the gains generalize across video-fMRI datasets.
Qualitative comparisons with ground-truth videos are in Fig.~\ref{fig:hcp_vis}.

\noindent\textbf{Image Reconstruction.}
We further evaluate \methodname{} on NSD for fMRI-to-image reconstruction.
As shown in Table~\ref{tab:unified_comparison}, \methodname{} achieves strong performance across both low-level and high-level metrics, obtaining the best average PixCorr score of 0.355 and competitive semantic scores on AlexNet, Inception, and CLIP-based evaluations.
These results show that treating images as single-frame visual streams preserves static image reconstruction quality while allowing the same visual-stream framework to support video reconstruction.

\subsection{Ablation on Main Extensions}

We ablate the main extensions introduced in \methodname{}: multi-subject image-fMRI pretraining, the Serial Visual Decoupler, and image-to-image diffusion refinement.

\noindent\textbf{Video-side ablation.}
We first evaluate the contribution of multi-subject pretraining and the Serial Visual Decoupler on video reconstruction.
As shown in Table~\ref{tab:ablation}, both extensions improve different aspects of video reconstruction over the conference version \confname{}.
Multi-subject pretraining mainly improves semantic retrieval, while the Serial Visual Decoupler brings larger gains in both semantic-level and spatiotemporal metrics.
Combining both extensions achieves the best overall performance, showing that image-fMRI pretraining and serial visual decoding are complementary.

\noindent\textbf{Image-side ablation.}
We further conduct an image-side ablation on NSD Subject 1 to examine whether these extensions also benefit static image reconstruction.
As shown in Table~\ref{tab:ablation_nsd}, removing multi-subject pretraining or the Serial Visual Decoupler degrades both low-level and high-level image reconstruction metrics, with the Serial Visual Decoupler causing the largest overall drop.
Removing the image-to-image diffusion refinement stage improves some semantic feature metrics but reduces low-level fidelity, suggesting a trade-off between visual structure and feature-level semantics.
Together, the video- and image-side ablations support the effectiveness of multi-subject pretraining and serial visual decoding across both dynamic and static visual-stream reconstruction.

\begin{figure}[t]
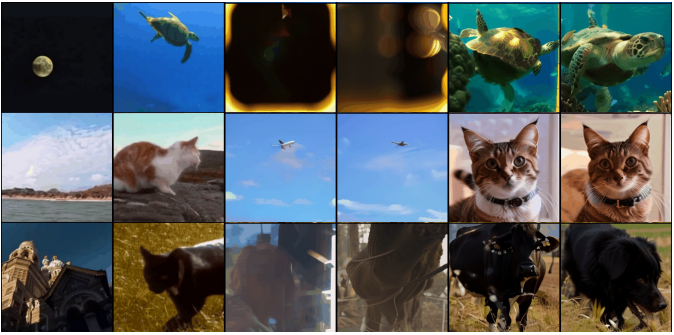

\centering
    \animategraphics[width=\linewidth]{8}{videos/delta_fMRI/video1/}{01}{06}
    \animategraphics[width=\linewidth]{8}{videos/delta_fMRI/video2/}{01}{06}
    \animategraphics[width=\linewidth]{8}{videos/delta_fMRI/video3/}{01}{06}
    \caption{%
    \textbf{Delta-fMRI counterfactual synthesis.}
    Each row shows reconstructed keyframes for one clip pair $(i,j)$ with six
    columns: the two ground-truth clips $(GT_i,GT_j)$ followed by
    reconstructions at $\alpha\in\{-0.5,-0.3,+0.3,+0.5\}$. Negative
    $\alpha$ values interpolate from $\mathbf{x}_j$ toward $\mathbf{x}_i$,
    whereas positive $\alpha$ values extrapolate beyond $\mathbf{x}_j$ along
    the $\mathbf{x}_i\rightarrow\mathbf{x}_j$ direction. The resulting visual
    changes are smooth and semantically coherent under graded fMRI
    perturbations.}
  \label{fig:delta_fmri_montage}
\end{figure}

\section{Beyond Reconstruction: Neural-to-Visual Interpretability}
\label{sec:neuro_analysis}

Beyond reconstruction fidelity, we conduct three analyses to examine whether
\methodname{} learns brain-grounded neural-to-visual representations. First,
delta-fMRI perturbation tests whether graded changes along a neural direction
produce direction-consistent visual changes. Second, ROI ablation tests whether
the encoder read-out depends on the cortical regions actually engaged by each
stimulus. Third, ROI amplification tests whether selectively enhancing
functional ROIs can steer the decoded semantics. The delta-fMRI analysis is
conducted on the CC2017 video-fMRI dataset, whereas ROI ablation and ROI
amplification are conducted on CC2017 subject~1. Additional mathematical
details, implementation notes, and cross-subject replications are provided in
Supplementary~\S\ref{app:neuro_supp}.

\subsection{\methodname{} Exhibits Directional Controllability under Delta-fMRI Perturbation}
\label{subsec:delta_fmri}

Reconstruction metrics evaluate whether the model recovers a fixed
ground-truth stimulus. A complementary property is controllability: when the
input fMRI response is moved along a meaningful neural direction, the decoded
visual content should change coherently along the same direction.

Let $\mathbf{x}_i$ and $\mathbf{x}_j$ denote the trial-averaged fMRI responses
to two CC2017 test clips. We define a directional delta-fMRI perturbation
around $\mathbf{x}_j$ as
\begin{equation}
    \tilde{\mathbf{x}}_j(\alpha)
    =
    \mathbf{x}_j
    +
    \alpha(\mathbf{x}_j-\mathbf{x}_i),
    \quad
    \alpha \in \{-0.5,-0.3,0.3,0.5\}.
\end{equation}
Negative $\alpha$ moves $\mathbf{x}_j$ toward $\mathbf{x}_i$, whereas positive
$\alpha$ extrapolates beyond $\mathbf{x}_j$ along the same neural direction.
Each synthetic response is z-score normalized before being passed to the
encoder. Because the normalization is performed per response, this experiment
is interpreted as a qualitative direction-consistency probe rather than a
metrically calibrated proportionality test; the normalization analysis is
provided in Supplementary~\S\ref{app:delta_fmri_details}.

As shown in Fig.~\ref{fig:delta_fmri_montage}, the reconstructed keyframes
change smoothly along the perturbation direction. Negative $\alpha$ values
produce reconstructions closer to clip $i$, while positive $\alpha$ values
move beyond clip $j$ along the same direction. This suggests that the encoder
propagates graded fMRI perturbations into coherent perceptual changes rather
than unstructured visual variation.

\begin{figure*}[t]
  \centering
  \includegraphics[width=\linewidth]{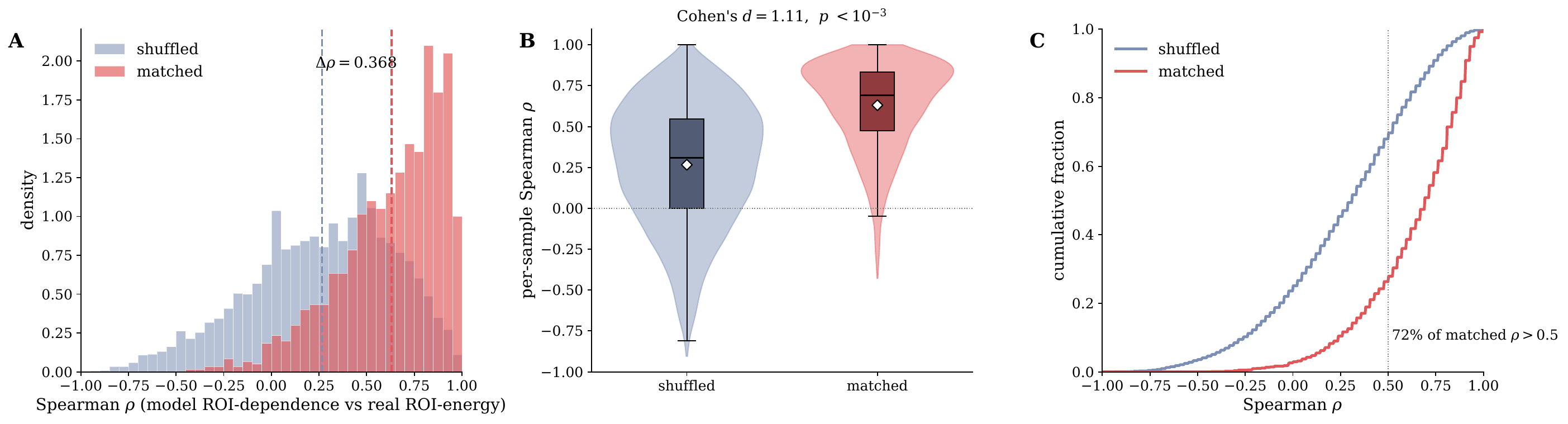}
  \caption{%
    \textbf{Per-sample stimulus specificity of ROI causal dependence.}
    For each test sample, the ROI ablation profile ($\Delta\cos$ across the
    eight ROIs) is correlated by Spearman's $\rho$ with the sample's own
    fMRI ROI activation-energy profile (\emph{matched}, red) or with that of
    a randomly paired sample (\emph{mismatched}, blue). A higher $\rho$
    indicates that the encoder relies most on the ROIs that the stimulus most
    strongly drives.}
  \label{fig:stimulus_specificity}
\end{figure*}

\begin{figure}[t]
  \centering
  \includegraphics[width=0.9\linewidth]{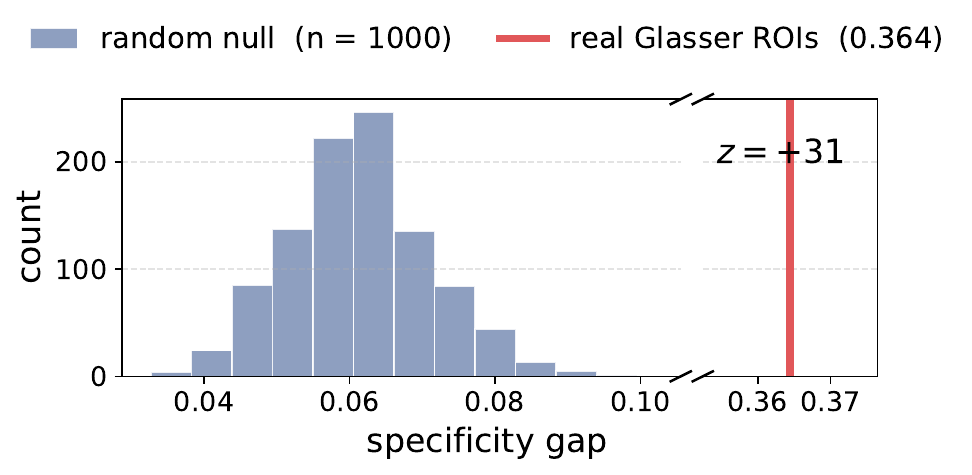}
  \caption{%
    \textbf{Random-grouping control for ROI ablation specificity.}
    The matched-minus-mismatched specificity gap obtained with the real
    Glasser parcellation is compared against a null distribution computed
    under size-matched random voxel groupings. The real ROI result falls far
    outside the random-grouping null, ruling out the possibility that larger
    ROIs trivially dominate the ablation signal.}
  \label{fig:roi_ablation_random_control}
\end{figure}

\subsection{\methodname{} Learns Stimulus-Relevant ROI Features through ROI Ablation}
\label{subsec:roi_ablation}

A visually accurate decoder may still ignore the spatial organization of
cortical responses. We therefore ask whether the encoder exhibits
stimulus-specific dependence on cortical regions, namely whether it relies
most strongly on the ROIs that each stimulus preferentially activates.

We assign the activated voxels of CC2017 subject~1 to eight functional ROIs
using the Glasser multi-modal parcellation~\cite{glasser2016multi}:
\textsc{face}, \textsc{place}, \textsc{motion}, \textsc{early\_visual},
\textsc{object}, \textsc{semantic}, \textsc{ventral}, and \textsc{dorsal}.
Among $V=13{,}447$ activated voxels, $6{,}164$ fall within these ROIs. The
full HCP-MMP1 parcel assignments are listed in
Supplementary~\S\ref{app:roi_definitions}. The first six ROIs are category- or
stage-selective and mutually disjoint, whereas \textsc{ventral} and
\textsc{dorsal} denote broader visual streams and overlap with finer parcels
by construction. Each ROI is nevertheless ablated independently.

For each test sample $\mathbf{x}_i$, we zero out one ROI at a time and measure
the resulting change in the encoder read-out. Let
$\mathbf{m}_r\in\{0,1\}^{V}$ denote the binary ablation mask for ROI $r$,
where $m_{r,k}=0$ if voxel $k\in\mathcal{R}_r$ and $m_{r,k}=1$ otherwise.
We define the ROI sensitivity score as
\begin{equation}
    \Delta \cos_r(\mathbf{x}_i)
    =
    1 -
    \cos\big(
    f(\mathbf{x}_i),
    f(\mathbf{x}_i \odot \mathbf{m}_r)
    \big),
\end{equation}
where $f=\mathcal{P}_i\circ\mathcal{B}_{\theta}\circ\mathrm{ridge}$ denotes
the image-level encoder read-out that produces the fMRI-derived CLIP visual
embedding. A larger $\Delta\cos_r$ indicates that ablating ROI $r$ causes a
larger change in the predicted embedding. We compare this sensitivity profile
with the corresponding fMRI activation-energy profile,
\begin{equation}
    E_r(\mathbf{x}_i)
    =
    |\mathcal{R}_r|^{-1}
    \sum_{k\in\mathcal{R}_r}x_{ik}^{2}.
\end{equation}
The embedding-level read-out is used to isolate encoder sensitivity from the
additional stochasticity of the downstream diffusion generator; a detailed
discussion is provided in Supplementary~\S\ref{app:roi_ablation_details}.

We compute the per-sample matched Spearman correlation between
$\Delta\cos_{\cdot}(\mathbf{x}_i)$ and $E_{\cdot}(\mathbf{x}_i)$, and assess
significance with a sample-level paired permutation test. The matched
distribution is strongly right-shifted
($\bar{\rho}_{\rm m}=0.631\pm0.27$, $n=1{,}200$) relative to the mismatched
baseline ($\bar{\rho}_{\rm mm}=0.268\pm0.38$), yielding a matched-minus-null
gap of $\Delta\rho=0.368$ with empirical $p<10^{-3}$ and pooled-SD
Cohen's $d=1.11$ (Fig.~\ref{fig:stimulus_specificity}). This suggests that
the encoder read-out is most sensitive to the ROIs that are most strongly
activated by the corresponding stimulus.

A remaining concern is that spatially contiguous or larger voxel groups may
cause stronger degradation regardless of functional identity. To address this,
we repeat the analysis on $1{,}000$ random size-matched partitions, where the
$6{,}164$ ROI-assigned voxels are randomly permuted and divided into eight
groups matching the real ROI sizes. The real Glasser parcellation produces a
much larger matched-minus-mismatched gap than the random null
($0.364$ vs.\ $0.061\pm0.010$), and no random partition reaches the real gap
(empirical $p<10^{-3}$; Fig.~\ref{fig:roi_ablation_random_control}). Thus, the
observed specificity cannot be explained by voxel energy or ROI size alone.

\begin{figure*}[t]
  \centering
  \includegraphics[width=\linewidth]{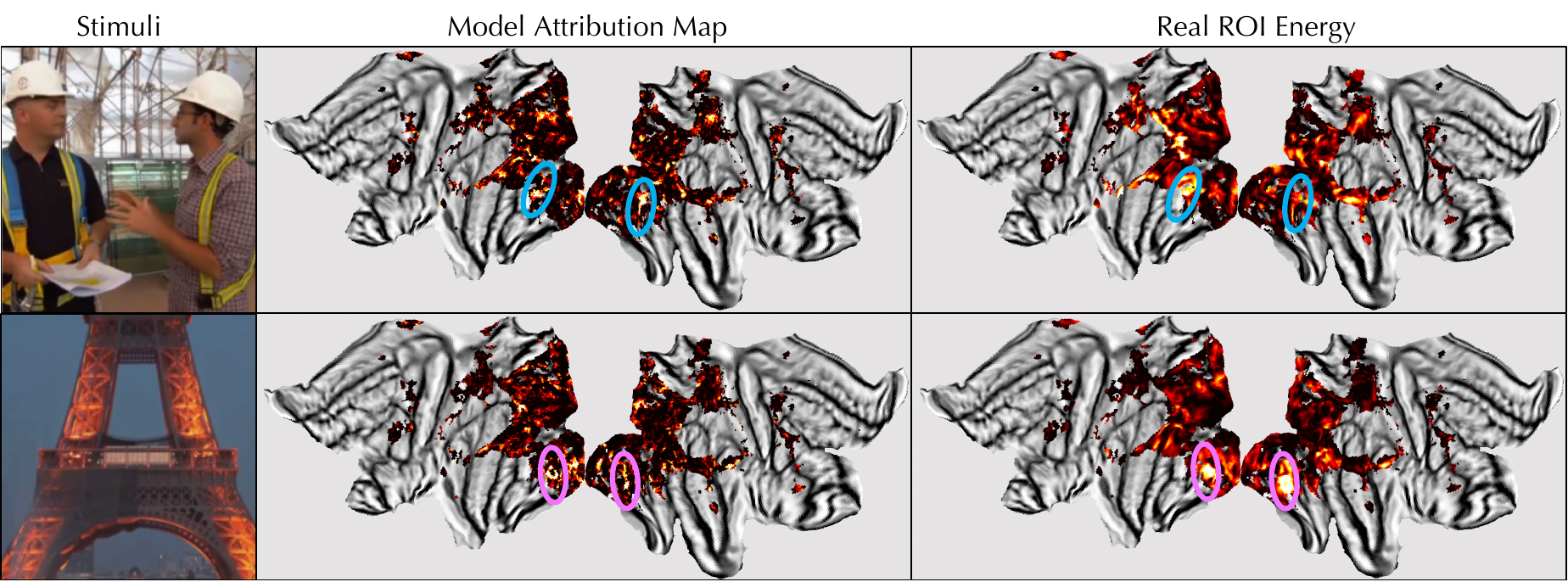}
  \caption{%
    \textbf{Model attribution maps match real cortical activation energy.}
    From left to right: the stimulus frame, the model attribution map projected
    onto the cortical surface, and the real fMRI ROI activation energy. For a
    face stimulus, the model highlights face-selective regions; for a scene
    stimulus, the attribution shifts toward place-selective regions.}
  \label{fig:attribution_vs_energy}
\end{figure*}

Beyond ROI-level ablation, we compute a voxel-level attribution map using the
positive input-gradient product
$\mathbf{a}(\mathbf{x})=\mathrm{ReLU}(\mathbf{x}\odot\nabla_{\mathbf{x}}
\lVert f(\mathbf{x})\rVert^2)$. As shown in
Fig.~\ref{fig:attribution_vs_energy}, the attribution is broadly distributed
but highlights stimulus-appropriate regions, including face-selective cortex
for face stimuli and place-selective cortex for scene stimuli. This
voxel-level result provides complementary evidence that the encoder uses
stimulus-relevant cortical features rather than a region-agnostic shortcut.

\begin{figure}[t]
  \centering
  \includegraphics[width=0.9\linewidth]{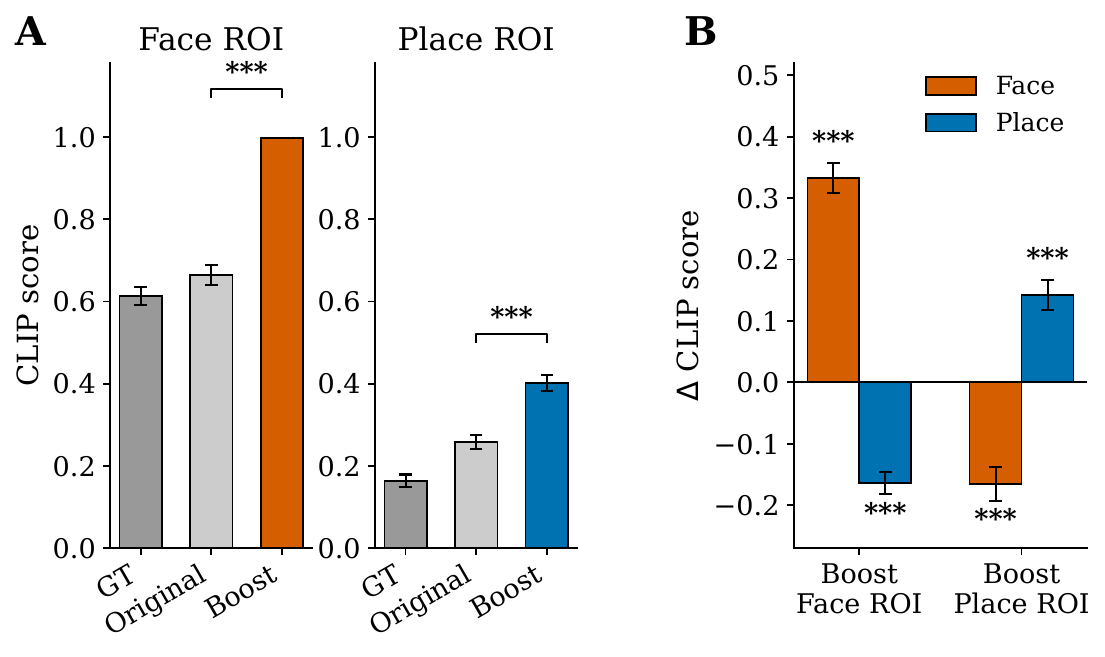}
  \caption{%
    \textbf{Matched semantic lift and asymmetric double dissociation under
    targeted ROI boosting.}
    (A) Bar plots show CLIP-face and CLIP-place scores before and after
    boosting the corresponding ROI. Each boost selectively elevates the
    semantic score matched to its known functional role.
    (B) Bars display the change ($\Delta$) in semantic scores when boosting
    the face ROI or the place ROI. Targeted boosting enhances its matched
    semantic score while suppressing the competing category's score.}
  \label{fig:boost_and_dissociation}
\end{figure}

We further replicate the ROI-ablation analysis across additional CC2017
subjects and all three HCP subjects. The per-sample matched-versus-null
specificity remains large and highly significant in every case
($\Delta\rho\in[0.33,0.46]$, pooled-SD Cohen's $d\in[0.86,1.27]$,
empirical $p<10^{-3}$ throughout); full per-subject results are reported in
Supplementary~\S\ref{app:additional_subjects}.

\subsection{\methodname{} Enables ROI-Guided Semantic Control through ROI Amplification}
\label{subsec:roi_enhance}

The ablation analysis establishes that the encoder reads stimulus-relevant
ROIs. We next ask the converse: whether writing to these ROIs by artificially
amplifying their responses before decoding can steer the reconstruction toward
the semantic content the regions are known to represent.

Given a target ROI $\mathcal{R}$ and the original trial-averaged response
$\mathbf{x}\in\mathbb{R}^V$ of CC2017 subject~1, we apply a boost
intervention: all voxels in $\mathcal{R}$ are set to a fixed high activation
value ($v=3.0$), non-target voxels are suppressed by a factor of $0.1$, and the
resulting response is z-score normalized before decoding. This is a deliberately
strong intervention, so we interpret the result as evidence of
region-selective, directionally specific semantic modulation, not as a
calibrated estimate of per-ROI causal gain. Additional details and limitations
are provided in Supplementary~\S\ref{app:roi_boost_details}.

\begin{table}[t]
\centering
\caption{Asymmetric double dissociation under ROI boost ($N{=}200$ samples,
  Wilcoxon signed-rank vs.\ original; $^{***}p{<}0.001$).}
\label{tab:roi_enhance_dissociation}
\resizebox{\columnwidth}{!}{%
\begin{tabular}{lcc}
\toprule
Boosted ROI & $\Delta$ CLIP-face & $\Delta$ CLIP-place \\
\midrule
Face (FFC + PeEc)          & $+0.332$ ($d{=}{+}1.13$)$^{***}$ & $-0.164$ ($d{=}{-}0.83$)$^{***}$ \\
Place (PHA1/2/3 + VMV1--3) & $-0.166$ ($d{=}{-}0.52$)$^{***}$ & $+0.143$ ($d{=}{+}0.52$)$^{***}$ \\
\bottomrule
\end{tabular}%
}
\end{table}

\begin{figure}[t]
  \centering
  \includegraphics[width=\linewidth]{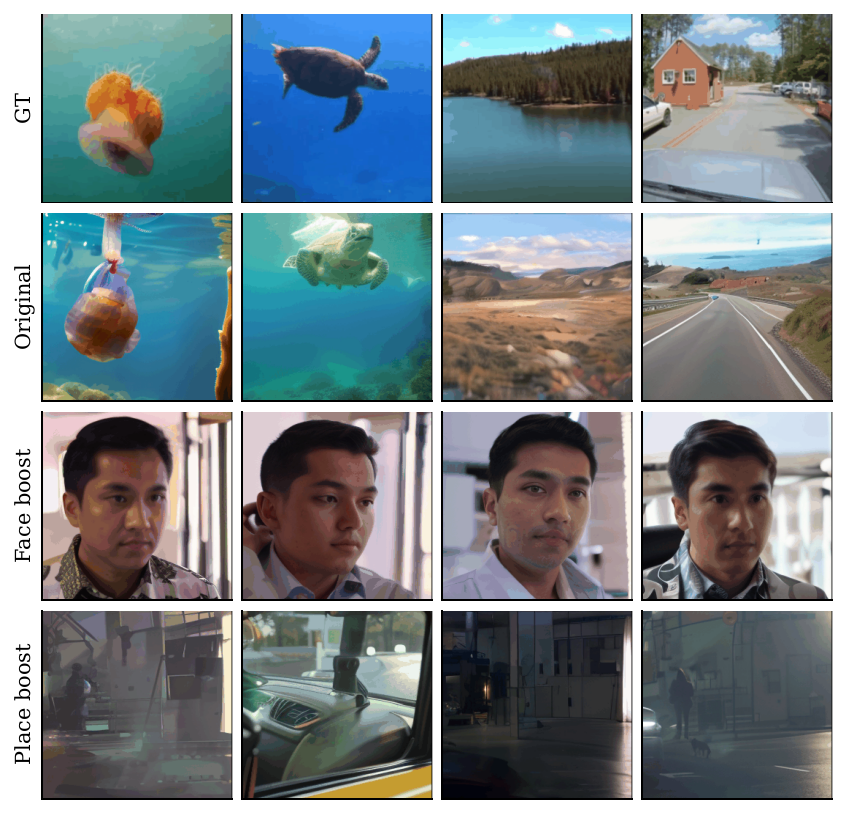}
  \caption{%
    \textbf{Boosting functional ROIs steers decoded content toward the matched
    category.}
    Each column shows one example stimulus; rows show the ground-truth frame,
    the unmanipulated reconstruction, the face-ROI-boosted reconstruction, and
    the place-ROI-boosted reconstruction.}
  \label{fig:qualitative_montage}
\end{figure}

We quantify face and place content using a zero-shot CLIP classifier, with
prompt sets and scoring formulas reported in Supplementary~\S\ref{app:clip_scoring}.
Boosting each ROI raises its matched semantic score
(Fig.~\ref{fig:boost_and_dissociation}, Table~\ref{tab:roi_enhance_dissociation}).
The face boost drives the CLIP-face score from $0.665$ to $0.997$
($\Delta={+}0.332$, $d={+}1.13$, $p<0.001$), indicating a strong shift toward
face-dominated reconstructions. The place boost increases the CLIP-place score
from $0.258$ to $0.401$ ($\Delta={+}0.143$, $d={+}0.52$, $p<0.001$), showing a
significant but weaker shift toward place content. Both boosts also suppress
the competing category: face boosting lowers CLIP-place
($\Delta={-}0.164$, $d={-}0.83$, $p<0.001$), while place boosting lowers
CLIP-face ($\Delta={-}0.166$, $d={-}0.52$, $p<0.001$). This asymmetric double
dissociation indicates that targeted ROI amplification exerts a
region-selective influence on decoded semantics, robustly for face and more
weakly for place.

Representative reconstructions are shown in Fig.~\ref{fig:qualitative_montage}.
We further replicate ROI boosting on all three HCP subjects and two additional
CC2017 subjects. The HCP subjects reproduce the asymmetric double dissociation
cleanly, whereas the additional CC2017 subjects show partial sign instability
under this deliberately out-of-distribution intervention
(Supplementary~\S\ref{app:additional_subjects}).
\section{Conclusion}

In this study, we present \methodname{}, a unified fMRI-to-visual-stream reconstruction framework for both static image and dynamic video reconstruction. 
By treating images as single-frame visual streams and videos as multi-frame visual streams, \methodname{} connects two previously separated decoding settings under a common formulation and architectural template. 
Compared with our conference version \textsc{Neurons}, \methodname{} extends the task paradigm from video-only reconstruction to visual-stream reconstruction for both images and videos, and replaces parallel task aggregation with serial cortex-inspired visual decoupling.
The proposed framework combines multi-subject image-fMRI model pretraining with subject-specific adaptation, allowing the spatial-semantic representations learned from image-fMRI data to benefit both static and dynamic visual-stream reconstruction. 
Through serial decoding, \methodname{} progressively organizes fMRI-derived representations into spatial, semantic, language-level, and temporal cues, which support high-fidelity generative reconstruction. 
Extensive experiments on image and video reconstruction benchmarks show that \methodname{} achieves state-of-the-art or competitive performance across low-level, semantic-level, and spatiotemporal metrics.
Beyond reconstruction accuracy, we further evaluate the neural-to-visual interpretability of \methodname{} through delta-fMRI perturbation, ROI ablation, and ROI amplification analyses. 
These analyses suggest that \methodname{} exhibits qualitative directional controllability under graded neural perturbation, that its encoder read-out is sensitive to stimulus-relevant cortical regions, and that it supports ROI-guided semantic control. 
Together, these results show that \methodname{} advances fMRI-based visual reconstruction toward a unified and interpretable visual-stream decoding framework.
{
\small
\bibliographystyle{IEEEtran}
\bibliography{main,mybib}
}

\clearpage
\setcounter{section}{0}
\setcounter{figure}{0}
\setcounter{table}{0}
\renewcommand\thesection{\Alph{section}}
\renewcommand\thesubsection{\thesection.\arabic{subsection}}
\renewcommand\thetable{\alph{table}}
\renewcommand\thefigure{\alph{figure}}
\renewcommand{\theHsection}{supp.\arabic{section}}
\renewcommand{\theHfigure}{supp.\arabic{figure}}
\renewcommand{\theHtable}{supp.\arabic{table}}

\twocolumn[
\begin{center}
    {\LARGE \bfseries Supplementary Material\par}
    \vspace{0.8em}
    {\Large \bfseries \methodname{}: A Unified Visual-Stream Framework\\ for Image and Video Reconstruction from fMRI \\
    with Neural-to-Visual Interpretability\par}
    \vspace{1.5em}
\end{center}
]

{
\par\noindent\rule{\linewidth}{0.4pt}
\noindent{\large\bfseries Contents}\par
\vspace{0.4em}
\begin{itemize}
  \setlength{\itemsep}{0.15em}
  \setlength{\parskip}{0pt}
  \item[A.] \hyperref[sec:supp_dtc]{Decoupled Task Construction}
  \item[B.] \hyperref[sec:supp_model_details]{Additional Model Details}
  \item[C.] \hyperref[sec:supp_training_config]{Training and Inference Configuration}
  \item[D.] \hyperref[sec:supp_preprocessing]{Dataset Details and Preprocessing}
  \item[E.] \hyperref[sec:supp_implementation]{Implementation Overview}
  \item[F.] \hyperref[sec:supp_metrics]{Evaluation Metrics}
  \item[G.] \hyperref[app:roi_definitions]{Functional ROI Parcel Assignments}
  \item[H.] \hyperref[app:neuro_supp]{Additional Details for Neural-to-Visual Interpretability}
  \item[I.] \hyperref[app:clip_scoring]{Zero-Shot CLIP Scoring of Face and Place Content}
  \item[J.] \hyperref[subsec:grpo_refinement]{Additional Ablation: GRPO-based Image-to-Image Refinement}
  \item[K.] \hyperref[app:additional_subjects]{Additional Subjects and Datasets for ROI Analyses}
\end{itemize}
\par\noindent\rule{\linewidth}{0.4pt}
\vspace{0.6em}
}

\section{Decoupled Task Construction}
\label{sec:supp_dtc}

\begin{figure*}[ht]
    \centering
    \includegraphics[width=1\linewidth]{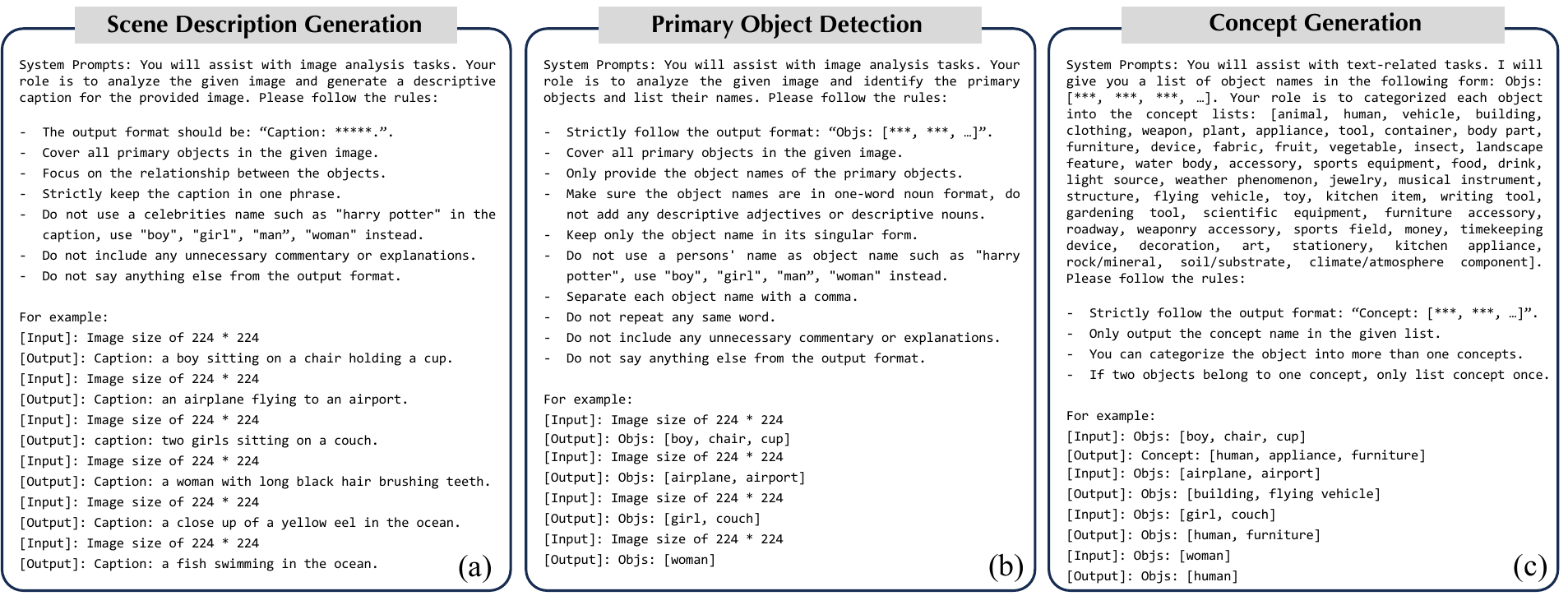}
    \caption{Instruction prompts for scene description generation and key concept generation.}
    \label{fig:prompts}
\end{figure*}

\begin{figure}[tb]
    \centering
    \includegraphics[width=1\linewidth]{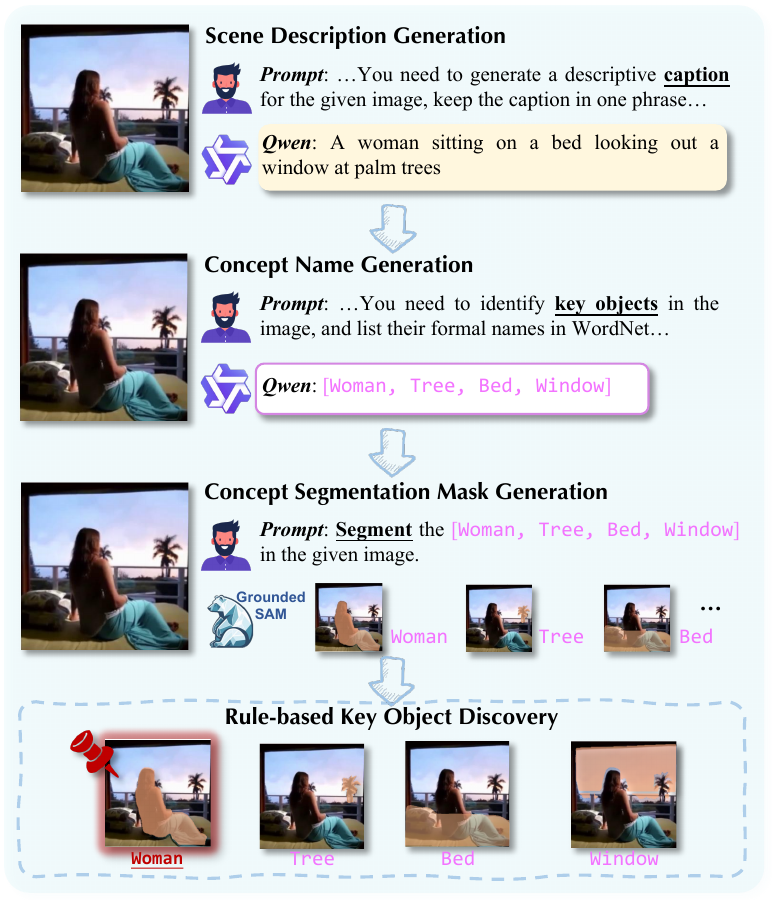}
    \caption{Construction pipeline for the decoupled task labels.}
    \label{fig:setup}
\end{figure}

We construct four types of task labels for the Serial Visual Decoupler: scene descriptions, concept names, concept segmentation masks, and rule-based key objects, as illustrated in Fig.~\ref{fig:setup}.

\noindent\textbf{Scene description generation.}
To construct labels for the scene description task, we generate a descriptive caption for each frame using Qwen2.5-VL-72B-Instruct~\cite{qwen}. 
Each frame is provided to Qwen with an instruction to produce a detailed one-phrase caption, such as ``\texttt{a woman sitting on a bed looking out a window at palm trees}''. 
The instruction prompts are shown in Fig.~\ref{fig:prompts}.

\noindent\textbf{Concept name generation.}
To construct labels for concept recognition, we generate concept names for the primary objects in each frame. 
We first input each frame into Qwen and ask it to recognize the primary objects, such as [\textquotesingle\texttt{woman}\textquotesingle, \textquotesingle\texttt{bed}\textquotesingle, \textquotesingle\texttt{tree}\textquotesingle]. 
The detected primary objects are then mapped into 51 concept categories summarized from WordNet~\cite{wordnet}, such as [\textquotesingle\texttt{animal}\textquotesingle, \textquotesingle\texttt{human}\textquotesingle, \textquotesingle\texttt{plant}\textquotesingle, \textquotesingle\texttt{furniture}\textquotesingle]. 
The output is a concept name list that covers the primary objects in the frame. 
The instruction prompts and the 51-concept list are shown in Fig.~\ref{fig:prompts}.

\noindent\textbf{Concept segmentation mask generation.}
We use the object list of each frame to generate segmentation masks for the primary objects. 
Each frame and its object list are provided to Grounded-SAM~\cite{ren2024groundedSAM}, which generates binary masks using the object names as textual prompts.

\noindent\textbf{Rule-based key object discovery.}
We identify key objects in video sequences by integrating motion dynamics, object size, and semantic importance. 
Background categories such as sky and ocean are excluded using predefined labels. 
For the remaining objects, inter-frame displacement is calculated, with higher weights assigned to priority categories such as humans and animals to reflect semantic importance. 
Objects are ranked by weighted displacement, and excessively large objects occupying more than 50\% of the image area are filtered out. 
Priority categories are selected first; otherwise, top-ranked objects by motion dynamics are chosen. 
If all objects are background, the largest object is selected as a fallback to ensure that at least one key object is identified.

\section{Additional Model Details}
\label{sec:supp_model_details}

\subsection{Brain Encoding and Diffusion-Prior Objective}
\label{sec:supp_BMdetail}

The brain encoding module follows the MindEye2-style design~\cite{mindeye2} and consists of a ridge regression module, a residual MLP module, and a diffusion prior network. 
The ridge regression module first maps the high-dimensional voxel response $\mathbf{x}_c$ into a lower-dimensional representation, and the residual MLP backbone $\mathcal{B}_{\theta}$ produces the neural representation $\mathbf{h}_c=\mathcal{B}_{\theta}(\mathrm{ridge}(\mathbf{x}_c))$. 
As described in Sec.~\ref{sec:brain_model}, the projection heads $\mathcal{P}_i$, $\mathcal{P}_{vid}$, and $\mathcal{P}_t$ then map $\mathbf{h}_c$ into the image-level visual tokens $\mathbf{e}^{i}_c$, the visual-stream embedding $\mathbf{e}^{v}_c$, and the textual embedding $\mathbf{e}^{t}_c$ used in the main paper; the image-level read-out probed in the ROI ablation (Sec.~\ref{subsec:roi_ablation}) is the composition $f=\mathcal{P}_i\circ\mathcal{B}_{\theta}\circ\mathrm{ridge}$. 
These embeddings are subsequently passed to the Serial Visual Decoupler in Sec.~\ref{sec:decoupler}, whose task-specific heads are denoted as $\mathcal{D}_{seg}$, $\mathcal{D}_{cls}$, $\mathcal{D}_{txt}$, and $\mathcal{D}_{rec}$. 
The Serial Visual Decoupler operates on $\mathbf{e}^{v}_c$ without gradient detachment, so the four decoupling losses backpropagate through $\mathcal{P}_{vid}$, $\mathcal{P}_i$, $\mathcal{B}_{\theta}$, and the ridge map; the read-out $f$ probed in Sec.~\ref{subsec:roi_ablation} is therefore the encoder of the fully trained NEURONS++ model, jointly shaped end-to-end by both the embedding-alignment objective $\mathcal{L}_{enc}$ and the cortex-inspired decoupling losses $\mathcal{L}_{dec}$ under Eq.~(\ref{eq:total}), rather than by CLIP alignment alone.
As defined in Eq.~(\ref{eq:enc}) of the main paper, the embedding-alignment objective $\mathcal{L}_{enc}$ comprises the visual CLIP alignment loss $\mathcal{L}_{\mathrm{CLIP}_v}$, the text CLIP alignment loss $\mathcal{L}_{\mathrm{CLIP}_t}$, and the diffusion-prior loss $\mathcal{L}_{prior}$, balanced by $\lambda_t$ and $\lambda_p$.
Here we specify these three terms in detail.
The diffusion prior is the component that contributes $\mathcal{L}_{prior}$; it is part of the brain encoding module rather than an additional task head in the Serial Visual Decoupler.
Consistent with the notation in Sec.~\ref{sec:brain_model}, $\mathcal{M}(\cdot)$ denotes the diffusion-prior \emph{module}, while $\epsilon_{\psi}(\cdot)$ denotes its internal noise-prediction network---the network actually optimized by $\mathcal{L}_{prior}$ below.

Following prior work, $\mathcal{L}_{\mathrm{CLIP}_v}$ and $\mathcal{L}_{\mathrm{CLIP}_t}$ are implemented with the BiMixCo objective, which combines MixCo augmentation and bidirectional contrastive learning. 
For each fMRI response $\mathbf{x}_c$, we randomly sample another response $\mathbf{x}_{m_c}$ and form a mixed response $\mathbf{x}^{*}_{c}=\lambda_c\mathbf{x}_c+(1-\lambda_c)\mathbf{x}_{m_c}$, where $\lambda_c$ is sampled from a Beta distribution. 
The mixed response is encoded and aligned with the corresponding visual or textual CLIP targets using the BiMixCo contrastive loss.

The diffusion prior is used to transform the fMRI-derived visual embedding into the CLIP visual-embedding space used by the generative prior. 
Let $\bar{\mathbf{e}}^{v}_c=\mathrm{Pool}(\mathbf{e}^{v}_c)$ denote the pooled fMRI-derived visual embedding, and let $\bar{\Phi}_v(\mathbf{Y}_c)=\mathrm{Pool}(\Phi_v(\mathbf{Y}_c))$ denote the pooled target CLIP visual embedding. 
For compactness, we write $\mathbf{e}_{x,c}=\bar{\mathbf{e}}^{v}_c$ and $\mathbf{e}_{y,c}=\bar{\Phi}_v(\mathbf{Y}_c)$. 
At diffusion step $s$, the noisy target embedding is
\[
    \tilde{\mathbf{e}}^{s}_{y,c}
    =
    \sqrt{\bar{\alpha}_s}\mathbf{e}_{y,c}
    +
    \sqrt{1-\bar{\alpha}_s}\boldsymbol{\epsilon},
    \quad
    \boldsymbol{\epsilon}\sim\mathcal{N}(\mathbf{0},\mathbf{I}),
\]
where $\bar{\alpha}_s$ is the cumulative noise schedule. 
The diffusion-prior module $\mathcal{M}(\cdot)$ is realized by a noise-prediction network $\epsilon_{\psi}(\cdot)$ that predicts the injected noise conditioned on the fMRI-derived embedding:
\begin{equation}
    \mathcal{L}_{prior}
    =
    \mathbb{E}_{c,\,s,\,\boldsymbol{\epsilon}}
    \left[
    \left\|
    \boldsymbol{\epsilon}
    -
    \epsilon_{\psi}
    \big(
    \tilde{\mathbf{e}}^{s}_{y,c},
    s,
    \mathbf{e}_{x,c}
    \big)
    \right\|_2^2
    \right].
\end{equation}
This formulation follows the standard conditional diffusion-prior training objective and makes explicit the roles of the fMRI-derived condition $\mathbf{e}_{x,c}$ and the target visual embedding $\mathbf{e}_{y,c}$.

\subsection{Serial Visual Decoupler Task Heads and Losses}
\label{sec:supp_decoupler_losses}
\label{sec:supp_seg}

The main paper describes the Serial Visual Decoupler at the architectural level, emphasizing the journal-version change from the parallel task aggregation of \confname{} to a progressive serial decoding process. Here we provide the task-head definitions and loss functions for the four inherited decoupled supervision targets: key object segmentation, concept recognition, scene description, and blurry visual-stream reconstruction. The construction of the corresponding labels follows \confname{} and is described in Supplementary~\S\ref{sec:supp_dtc}.

\noindent\textbf{Key object segmentation.}
The first stage focuses on low-level object structure. Given the stage-0 representation $\mathbf{z}^{0}_c$ and the text embedding of the detected key object, a text-driven segmentation decoder $\mathcal{D}_{seg}(\cdot)$ predicts the binary key object mask:
\begin{equation}
\label{eq:seg_pred_supp}
    \hat{\mathbf{M}}_c
    =
    \mathcal{D}_{seg}(\mathbf{z}^{0}_c,\mathbf{e}^{obj}_c),
    \quad
    \mathbf{e}^{obj}_c=\Phi_t(o_c),
\end{equation}
where $o_c$ is the detected key object name obtained during decoupled task construction and $\Phi_t(\cdot)$ denotes the CLIP text encoder. The compact segmentation objective is
\begin{equation}
\label{eq:seg}
    \mathcal{L}_{seg}
    =
    \mathcal{L}_{bce}(\hat{\mathbf{M}}_c,\mathbf{M}_c).
\end{equation}
For completeness, let $\mathbf{M}_c\in\{0,1\}^{T_c\times H\times W}$ denote the ground-truth mask of a single stream and $\hat{\mathbf{M}}_c\in[0,1]^{T_c\times H\times W}$ denote the predicted mask. Following the batch-index convention of Sec.~\ref{sec:problem}, the $b$-th stream of a minibatch is written $\mathbf{M}^{(b)}\equiv\mathbf{M}_{c=b}$ with entries $y^{seg,(b)}_{t,h,w}$, and likewise $\hat{y}^{seg,(b)}_{t,h,w}$ for the prediction. The binary cross-entropy loss is averaged over the minibatch, frames, and spatial locations:
\begin{equation}
\begin{aligned}
    \mathcal{L}_{seg}
    ={}&
    -\frac{1}{B T_c H W}
    \sum_{b=1}^{B}
    \sum_{t=1}^{T_c}
    \sum_{h=1}^{H}
    \sum_{w=1}^{W}
    \big[
    y^{seg,(b)}_{t,h,w}
    \log \hat{y}^{seg,(b)}_{t,h,w}
    \\
    &+
    (1-y^{seg,(b)}_{t,h,w})
    \log(1-\hat{y}^{seg,(b)}_{t,h,w})
    \big].
\end{aligned}
\end{equation}
This task provides spatial guidance for object localization and shape refinement.

\noindent\textbf{Concept recognition.}
The second stage learns object-level semantic concepts. We apply a multi-label classifier $\mathcal{D}_{cls}(\cdot)$ to the pooled representation of $\mathbf{z}^{1}_c$:
\begin{equation}
\label{eq:cls_pred_supp}
    \hat{\mathbf{C}}_c
    =
    \mathcal{D}_{cls}(\mathrm{Pool}(\mathbf{z}^{1}_c)),
\end{equation}
where $\mathrm{Pool}(\cdot)$ is the temporal- and token-axis average defined in Eq.~(\ref{eq:pool}) of the main paper, mapping $\mathbb{R}^{T_c\times N_{\mathrm{tok}}\times D}$ to $\mathbb{R}^{D}$. The concept recognition objective is a multi-label binary cross-entropy loss:
\begin{equation}
\label{eq:cls}
    \mathcal{L}_{cls}
    =
    \mathcal{L}_{bce}(\hat{\mathbf{C}}_c,\mathbf{C}_c).
\end{equation}
This stage constrains the reconstruction with object-level semantics and reduces conceptual shifts.

\noindent\textbf{Scene description.}
The third stage decodes scene-level language information. We use a text decoder $\mathcal{D}_{txt}(\cdot)$, implemented with GPT-2~\cite{gpt2}, to generate a caption conditioned on $\mathbf{z}^{2}_c$ and the fMRI-derived text embedding $\mathbf{e}^{t}_c$. Given the ground-truth token sequence $\mathbf{S}_c=\{s_1,\ldots,s_{|\mathbf{S}_c|}\}$, the language-modeling objective is
\begin{equation}
\label{eq:txt_supp}
    \mathcal{L}_{txt}
    =
    -\frac{1}{|\mathbf{S}_c|}
    \sum_{i=1}^{|\mathbf{S}_c|}
    \log p_{\mathcal{D}_{txt}}
    (s_i \mid s_{<i},\mathbf{z}^{2}_c,\mathbf{e}^{t}_c).
\end{equation}
This task provides language-level conditioning for downstream generation and helps preserve scene-level semantics.

\noindent\textbf{Blurry visual-stream reconstruction.}
The final stage learns coarse visual appearance and temporal structure. We use a reconstruction decoder $\mathcal{D}_{rec}(\cdot)$ to predict a low-resolution VAE-space visual stream:
\begin{equation}
\label{eq:rec_pred_supp}
    \hat{\mathbf{R}}_c
    =
    \mathcal{D}_{rec}(\mathbf{z}^{3}_c,\mathbf{e}^{t}_c),
    \quad
    \hat{\mathbf{R}}_c\in\mathbb{R}^{T_c\times C\times H'\times W'},
\end{equation}
where $H'=H/8$ and $W'=W/8$. Let $\mathbf{R}_c=\mathrm{VAE}(\mathbf{Y}_c)$ denote the VAE latent target of the visual stream. We optimize this task with a mean-absolute-error reconstruction loss, fully averaged over the batch, temporal, channel, and spatial dimensions:
\begin{equation}
\label{eq:rec_supp}
\begin{aligned}
    \mathcal{L}_{rec}
    ={}&
    \frac{1}{B T_c C H' W'}
    \sum_{b=1}^{B}
    \sum_{t=1}^{T_c}
    \sum_{\gamma=1}^{C}
    \sum_{h=1}^{H'}
    \sum_{w=1}^{W'}
    \\
    &\quad
    \left|
    \hat{R}^{(b)}_{t,\gamma,h,w}
    -
    R^{(b)}_{t,\gamma,h,w}
    \right|,
\end{aligned}
\end{equation}
where $\hat{R}^{(b)}_{t,\gamma,h,w}$ and $R^{(b)}_{t,\gamma,h,w}$ denote the predicted and target VAE-latent values of the $b$-th stream in the minibatch at frame $t$, channel $\gamma$, and spatial location $(h,w)$, with $\hat{\mathbf{R}}^{(b)}\equiv\hat{\mathbf{R}}_{c=b}$. For image reconstruction, this loss is applied to a single frame. For video reconstruction, it captures coarse temporal dynamics and color distribution across clips.

\noindent\textbf{Overall decoupling objective.}
The overall decoupling objective sums the four task losses with equal weight:
\begin{equation}
\label{eq:dec_supp}
    \mathcal{L}_{dec}
    =
    \mathcal{L}_{seg}
    +
    \mathcal{L}_{cls}
    +
    \mathcal{L}_{txt}
    +
    \mathcal{L}_{rec}.
\end{equation}
Each term is mean-reduced over its own output dimensions---frames and spatial locations for $\mathcal{L}_{seg}$, the concept set for $\mathcal{L}_{cls}$, the caption length $|\mathbf{S}_c|$ for $\mathcal{L}_{txt}$, and $(T_c,C,H',W')$ for $\mathcal{L}_{rec}$---so the four losses are of comparable scale rather than differing by orders of magnitude. We therefore combine them with equal weight and rely on the single outer coefficient $\lambda_{dec}$ in Eq.~(\ref{eq:total}) of the main paper to balance $\mathcal{L}_{dec}$ against the embedding-alignment objective $\mathcal{L}_{enc}$. The loss weights are reported in Supplementary~\S\ref{sec:supp_training_config}.

\section{Training and Inference Configuration}
\label{sec:supp_training_config}

For the CC2017 dataset, videos are downsampled from 30 FPS to 3 FPS to ensure fair comparison with prior methods. 
The blurry visual-stream prediction is interpolated to 8 FPS during inference to generate the final reconstructed video. 
The brain encoding module and Serial Visual Decoupler are trained for 30 and 50 epochs, respectively. 
The batch size is set to 120 for the brain encoding module and 10 for the Serial Visual Decoupler.

We use AdamW~\cite{loshchilov2017adamw} for optimization with a learning rate of $5\times10^{-5}$ and a OneCycle learning-rate schedule~\cite{smith2019onecircle}. 
The loss-balancing weights in Eqs.~(\ref{eq:enc}) and~(\ref{eq:total}) of the main paper are set to $\lambda_t = 0.5$, $\lambda_p = 1.0$, and $\lambda_{dec} = 1.0$ in all experiments.
For video reconstruction, the temporal expansion module can in principle be instantiated with different text-to-video diffusion models. 
In our implementation, we use the open-source AnimateDiff model~\cite{guo2023animatediff}, following prior fMRI-to-video reconstruction work~\cite{gong2025neuroclips}. 
Inference is performed with 25 DDIM steps~\cite{song2020ddim}.

For image reconstruction, the keyframe is first generated by an SDXL unCLIP pipeline and then refined by an SDXL-based image-to-image diffusion model.
The refinement uses the publicly available ZavyChromaXL~v3.0 SDXL checkpoint, conditioned on the fMRI-derived CLIP visual embedding and the decoded scene description.
The img2img process runs for 38 sampler steps and starts from a partially noised latent of the initial keyframe (i.e., the diffusion trajectory is entered near the high-noise end rather than from pure noise), so that the refined image preserves the layout of the keyframe while sharpening object-level detail.
All experiments are conducted on a single NVIDIA A100 GPU.

\section{Dataset Details and Preprocessing}
\label{sec:supp_preprocessing}

\subsection{CC2017}
\label{sec:supp_cc2017}

We use the public CC2017 video-fMRI dataset~\cite{wen2018neural} for fMRI-to-video reconstruction. 
The dataset contains fMRI responses from three subjects viewing natural movie clips subtending approximately $23^{\circ}\times23^{\circ}$ of visual angle, with a central fixation cross of $0.8^{\circ}\times0.8^{\circ}$. 
The stimuli are divided into 18 training movies and 5 testing movies, each 8 minutes long. 
The training and testing movies were presented 2 and 10 times to each subject, respectively. 
Anatomical MRI, including T1- and T2-weighted scans, and fMRI data were acquired using a 3T MRI system. 
The fMRI signals have a temporal resolution of 2 seconds.

We follow the preprocessing protocol used in prior fMRI-to-video reconstruction studies~\cite{wen2018neural,gong2025neuroclips}, based on the minimal preprocessing pipeline~\cite{glasser2013minimal}. 
The fMRI volumes are corrected for artifacts and head motion, registered to the MNI space, and projected onto cortical surfaces co-registered to a template~\cite{glasser2016multi}. 
To identify stimulus-activated voxels, we compute the voxel-wise correlation between the fMRI signals of the repeated presentations of each training movie. 
The correlation coefficients are Fisher $z$-transformed, and the averaged $z$-scores across the 18 training movie segments are evaluated with a one-sample $t$-test. 
Voxels with significant activation (Bonferroni-corrected, $P<0.05$) are selected to form a subject-specific visual mask, yielding $13{,}447$, $14{,}828$, and $9{,}114$ activated voxels in the visual cortex for the three subjects, respectively. 
This mask is computed only from the training set and then applied to both training and test responses. 
Following prior work~\cite{nishimoto2011reconstructing,han2019vaevideo}, we use BOLD signals with a 4-second hemodynamic lag to represent the corresponding movie stimulus responses.
After preprocessing, the dataset contains 8,640 training fMRI-video pairs and 1,200 testing pairs for each subject. 
The testing responses are averaged across repeated presentations, following the standard evaluation protocol.

\subsection{HCP}
\label{sec:supp_hcp}

We further evaluate fMRI-to-video reconstruction on the movie-watching data from the Human Connectome Project (HCP)~\cite{marcus2011informatics}. 
The full dataset contains BOLD responses from 158 subjects. 
In our experiments, we use three subjects, 100610, 102816, and 104416, following the setting of prior video reconstruction work. 
The data were acquired using a 7T MRI scanner with a spatial resolution of 1.6 mm and a repetition time of 1 second.

We use the standard HCP preprocessing pipeline, including head-motion correction, distortion correction, high-pass filtering, and removal of temporal artifacts using independent component analysis (ICA). 
The preprocessed BOLD responses are registered to the MNI standard space. 
Since the HCP movie-watching data do not provide repeated presentations in the same way as CC2017, we identify visually responsive voxels using a split-half reliability test on the cortical surface ($59{,}412$ fsLR-32k vertices).
For each movie, we divide its clips into even- and odd-indexed halves and compute the voxel-wise Pearson correlation between the two halves within consecutive segments; the correlations are Fisher $z$-transformed and tested against zero with a per-voxel one-sample $t$-test, and voxels surviving Bonferroni correction ($\alpha=0.01$) are retained. If fewer voxels survive than the target budget, we fall back to the top $12{,}000$ voxels ranked by adjusted $p$-value. This yields $12{,}000$ voxels for subjects 100610 and 102816, and $13{,}292$ for subject 104416.

Following prior fMRI video decoding protocols~\cite{nishimoto2011reconstructing,han2019vaevideo,wang2022reconstructing}, we use a 4-second hemodynamic lag to align BOLD responses with movie stimuli.
Since there is no widely established train/test split for video reconstruction on these three HCP subjects, the split must be constructed by us.
The HCP movie-watching stimuli are continuous films, and the clips are 1-second segments cut sequentially from four movies (728, 792, 745, and 775 clips, respectively; 3{,}040 in total).
A naive random shuffle of all clips into a 90\%/10\% split (as used in the conference version) causes severe temporal leakage: because temporally adjacent clips are highly correlated and the 4-second hemodynamic lag spreads each clip's BOLD response onto its neighbors, a randomly held-out test clip is almost always flanked by training clips that carry the same hemodynamic signal.
Concretely, under the random split, $100\%$ of test clips have a training clip within the 4-second hemodynamic window and $98\%$ have an immediate ($\pm1$\,s) training neighbor (median nearest-training gap of 1\,s), and the voxel-wise fMRI response of a test clip correlates with its adjacent training clip at $r{=}0.43$ on average, which inflates test-time scores.

To eliminate this leakage, we adopt a \emph{group-aware, embargoed} split.
Within each movie, we partition the timeline into contiguous 30-second blocks and assign whole blocks to the training or test set (so all four movies are represented in both sets), and at every train/test boundary we discard an embargo of 4 seconds (the hemodynamic lag) of training clips adjacent to any test clip.
This yields 2{,}597 training and 355 test clips (88 boundary clips discarded by the embargo) and reduces the fraction of test clips with a training neighbor within the hemodynamic window from $100\%$ to $0\%$ (minimum nearest-training gap of 5\,s, median 13\,s), removing the temporal leakage.
All HCP results in this paper, including those in Table~\ref{tab:unified_comparison}, are reported under this leakage-free split, and the split is fixed across methods to ensure fair comparison.

\subsection{Natural Scenes Dataset}
\label{sec:supp_nsd}

For fMRI-to-image reconstruction, we use the Natural Scenes Dataset (NSD)~\cite{NSD}, a large-scale public fMRI dataset containing brain responses to natural images from COCO~\cite{lin2014microsoft}. 
The dataset includes 8 subjects, each scanned for 30--40 hours across 30--40 scanning sessions. 
Each scanning session contains 750 images, and each image is presented for 3 seconds. 
Most images are unique to each subject, while a shared set of 1,000 images is presented to all subjects.

We follow the standard train/test split used in previous NSD reconstruction studies~\cite{takagi2023high,mindeye2,gong2025mindtuner}. 
The 1,000 shared images are used as the test set, and the remaining subject-specific images are used for training. 
We evaluate performance on the four subjects who completed all 40 scanning sessions. 
For the test set, responses to repeated presentations of the same image are averaged, resulting in 1,000 test samples per subject. 
For the training set, we use single-trial, non-averaged brain-image pairs, resulting in approximately 30,000 training samples per subject.

The fMRI responses are normalized single-trial beta estimates produced by GLMSingle~\cite{NSD}. 
We use preprocessed flattened fMRI voxels in 1.8-mm native volume space from the ``nsdgeneral'' brain region, which is defined by the NSD authors as the subset of voxels in posterior cortex most responsive to the presented visual stimuli. 
This region contains approximately 13,000 to 16,000 voxels per participant. 
For each subject, voxel-wise $z$-score normalization is applied using the mean and standard deviation computed from the training set only, and the same statistics are then used to normalize the test responses. 
We keep the shared 1,000-image test set fixed across all experiments to ensure consistency with prior fMRI-to-image reconstruction benchmarks.

\section{Additional fMRI-to-Video Qualitative Results}
\label{sec:additional_qualitative_results}

This section provides additional qualitative examples to complement the visual results reported in the main paper. We include more cases from both the HCP and CC2017 datasets to show the consistency of the reconstructed dynamic visual content across different subjects and stimuli.

\subsection{HCP}
\label{sec:additional_hcp}

\begin{figure}[t]
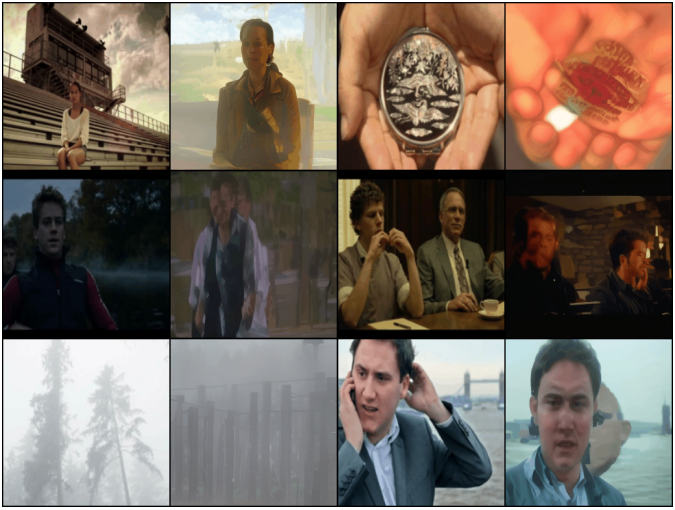

    \centering
    \animategraphics[width=0.5\linewidth]{8}{videos/hcp_2/034/}{01}{08}%
    \animategraphics[width=0.5\linewidth]{8}{videos/hcp_2/068/}{01}{08}\\

    \animategraphics[width=0.5\linewidth]{8}{videos/hcp_2/117/}{01}{08}%
    \animategraphics[width=0.5\linewidth]{8}{videos/hcp_2/129/}{01}{08}\\

    \animategraphics[width=0.5\linewidth]{8}{videos/hcp_2/203/}{01}{08}%
    \animategraphics[width=0.5\linewidth]{8}{videos/hcp_2/247/}{01}{08}\\
    \caption{Qualitative comparison of fMRI-to-video reconstruction on the HCP subject 2. For each example, the left clip/frame is the ground-truth stimulus and the right clip/frame is the reconstructed result; every row shows two examples. \textit{Best viewed with Acrobat Reader. Click the images to play the animation clips.}}
    \label{fig:hcp_2_vis}
\end{figure}

\begin{figure}[t]
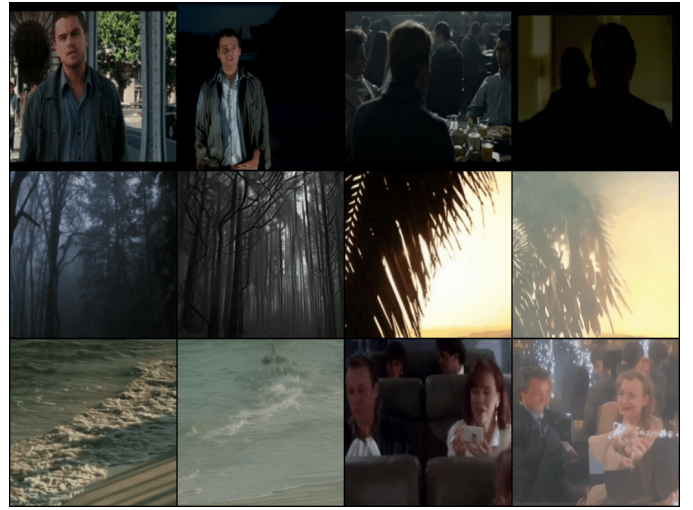

    \centering
    \animategraphics[width=0.5\linewidth]{8}{videos/hcp_3/096/}{01}{08}%
    \animategraphics[width=0.5\linewidth]{8}{videos/hcp_3/124/}{01}{08}\\

    \animategraphics[width=0.5\linewidth]{8}{videos/hcp_3/216/}{01}{08}%
    \animategraphics[width=0.5\linewidth]{8}{videos/hcp_3/219/}{01}{08}\\

    \animategraphics[width=0.5\linewidth]{8}{videos/hcp_3/225/}{01}{08}%
    \animategraphics[width=0.5\linewidth]{8}{videos/hcp_3/289/}{01}{08}\\
    \caption{Qualitative comparison of fMRI-to-video reconstruction on the HCP subject 3. For each example, the left clip/frame is the ground-truth stimulus and the right clip/frame is the reconstructed result; every row shows two examples. \textit{Best viewed with Acrobat Reader. Click the images to play the animation clips.}}
    \label{fig:hcp_3_vis}
\end{figure}

For the HCP dataset, we present supplementary results from subjects 2 and 3 in Fig.~\ref{fig:hcp_2_vis} and Fig.~\ref{fig:hcp_3_vis}. These additional examples further corroborate the effectiveness of our method across different subjects.

\subsection{CC2017}
\label{sec:additional_cc2017}

\begin{figure}[t]
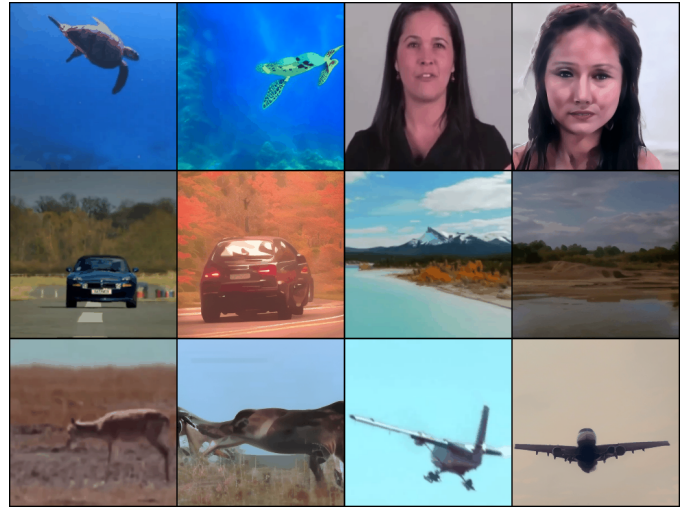

    \centering
    \animategraphics[width=0.5\linewidth]{8}{videos/cc2017_2/105/}{01}{08}%
    \animategraphics[width=0.5\linewidth]{8}{videos/cc2017_2/142/}{01}{08}\\

    \animategraphics[width=0.5\linewidth]{8}{videos/cc2017_2/388/}{01}{08}%
    \animategraphics[width=0.5\linewidth]{8}{videos/cc2017_2/408/}{01}{08}\\

    \animategraphics[width=0.5\linewidth]{8}{videos/cc2017_2/778/}{01}{08}%
    \animategraphics[width=0.5\linewidth]{8}{videos/cc2017_2/779/}{01}{08}\\
    \caption{Qualitative comparison of fMRI-to-video reconstruction on the CC2017 subject 2. For each example, the left clip/frame is the ground-truth stimulus and the right clip/frame is the reconstructed result; every row shows two examples. \textit{Best viewed with Acrobat Reader. Click the images to play the animation clips.}}
    \label{fig:cc2017_2_vis}
\end{figure}

\begin{figure}[t]
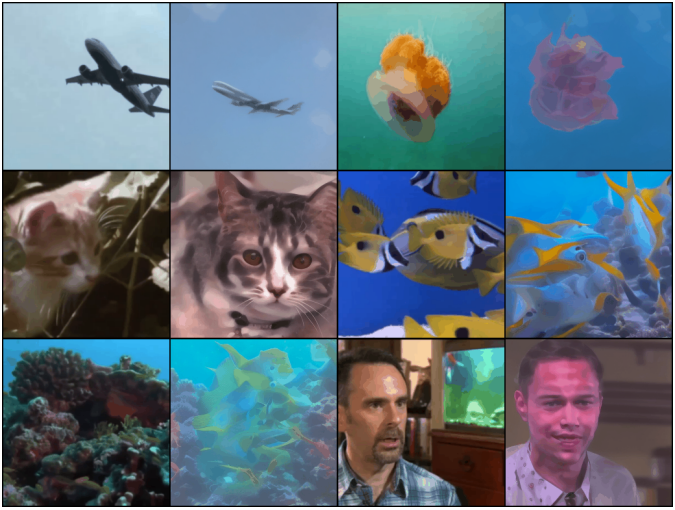

    \centering
    \animategraphics[width=0.5\linewidth]{8}{videos/cc2017_3/026/}{01}{08}%
    \animategraphics[width=0.5\linewidth]{8}{videos/cc2017_3/096/}{01}{08}\\

    \animategraphics[width=0.5\linewidth]{8}{videos/cc2017_3/293/}{01}{08}%
    \animategraphics[width=0.5\linewidth]{8}{videos/cc2017_3/403/}{01}{08}\\

    \animategraphics[width=0.5\linewidth]{8}{videos/cc2017_3/469/}{01}{08}%
    \animategraphics[width=0.5\linewidth]{8}{videos/cc2017_3/1195/}{01}{08}\\
    \caption{Qualitative comparison of fMRI-to-video reconstruction on the CC2017 subject 3. For each example, the left clip/frame is the ground-truth stimulus and the right clip/frame is the reconstructed result; every row shows two examples. \textit{Best viewed with Acrobat Reader. Click the images to play the animation clips.}}
    \label{fig:cc2017_3_vis}
\end{figure}

We further provide qualitative examples from CC2017 subjects 2 and 3 in Fig.~\ref{fig:cc2017_2_vis} and Fig.~\ref{fig:cc2017_3_vis}. These results offer an additional view of the model's reconstruction behavior on natural videos with diverse visual content.

\section{Implementation Overview}
\label{sec:supp_implementation}

We implement \methodname{} using the same preprocessing and evaluation protocols as prior fMRI-to-image and fMRI-to-video reconstruction studies to ensure fair comparison. 
For video reconstruction, each video is divided into short clips according to the temporal resolution of fMRI, and the corresponding BOLD responses are aligned with visual stimuli using a hemodynamic lag. 
For image reconstruction, each static image is treated as a single-frame visual stream, allowing the same visual-stream formulation to be applied to both image and video reconstruction.

All fMRI responses are normalized following the dataset-specific preprocessing protocols described in Supplementary~\S\ref{sec:supp_preprocessing}. 
The brain encoding module maps fMRI voxels into visual and textual embeddings. 
These embeddings are then processed by the Serial Visual Decoupler, whose four task heads correspond to $\mathcal{D}_{seg}$, $\mathcal{D}_{cls}$, $\mathcal{D}_{txt}$, and $\mathcal{D}_{rec}$ in Sec.~\ref{sec:decoupler}. 
During inference, image reconstruction uses the decoded cues for single-frame visual-stream generation, while video reconstruction further performs temporal expansion using the reconstructed keyframe, decoded scene description, key object mask, concept predictions, and blurry visual-stream reconstruction. 
Additional model details are provided in Supplementary~\S\ref{sec:supp_model_details}.

\section{Evaluation Metrics}
\label{sec:supp_metrics}

\subsection{Video Reconstruction Metrics}

For fMRI-to-video reconstruction, we evaluate the generated videos at both the semantic level and the spatiotemporal level.

\noindent\textbf{Semantic-level evaluation.}
Following prior fMRI-to-video reconstruction studies, we use an $N$-way classification-based evaluation protocol to measure whether the reconstructed video preserves the semantic category of the ground-truth stimulus.
Here $N$ denotes the number of candidate classes used in the metric and is unrelated to the CLIP visual-token count $N_{\mathrm{tok}}$ in the main paper.
Specifically, we use a video classifier based on VideoMAE~\cite{tong2022videomae}, trained on the Kinetics-400 dataset~\cite{kay2017kinetics}. 
For each test sample, the ground-truth video and reconstructed video are classified, and the trial is considered successful if the ground-truth class is correctly identified among $N$ randomly sampled candidate classes, including the true class. 
We report 2-way and 50-way accuracy following prior work.

\noindent\textbf{Spatiotemporal consistency.}
To evaluate temporal smoothness and consistency, we compute CLIP image embeddings for each frame of the reconstructed video and report the average cosine similarity between adjacent frames. 
This metric, denoted as CLIP-pcc, is commonly used in video editing and video reconstruction tasks~\cite{clippcc}. 
A higher CLIP-pcc indicates stronger temporal consistency between neighboring frames.

\subsection{Frame-Level and Image Reconstruction Metrics}

In addition to video-level metrics, we also evaluate reconstructed frames and static image reconstructions using both low-level and high-level metrics.

\noindent\textbf{Low-level metrics.}
We use SSIM and PSNR to evaluate pixel-level similarity between reconstructed frames/images and the corresponding ground-truth visual stimuli. 
These metrics measure low-level spatial fidelity, including structure, luminance, and pixel-wise reconstruction quality.

\noindent\textbf{High-level semantic metrics.}
For semantic-level image evaluation, we use an $N$-way top-$K$ classification test based on an ImageNet classifier with 1,000 classes. 
For each trial, the classifier predicts category probabilities for the reconstructed image. 
The trial is considered successful if the ground-truth class is included in the top-$K$ predicted probabilities among $N$ randomly sampled candidate classes, including the true class. 
We use top-1 prediction and repeat the random sampling procedure 100 times to obtain a stable success rate.

For NSD-based fMRI-to-image reconstruction, we additionally follow the standard evaluation protocol used in prior image reconstruction studies. 
The metrics include low-level similarity measures such as pixel correlation and SSIM, as well as high-level feature-based metrics computed with pretrained visual recognition networks, including AlexNet, Inception, CLIP, EfficientNet, and SwAV. 
These metrics jointly assess whether the reconstructed image preserves both low-level visual structure and high-level semantic content.

\begin{table*}[!t]
  \centering
  \caption{
  Full quantitative comparison of fMRI-to-image reconstruction on NSD.
  Results are averaged over subjects 1, 2, 5, and 7 from the Natural Scenes Dataset.
  N/A indicates that the method is video-only and does not support static image reconstruction.
  }
  \resizebox{0.9\linewidth}{!}{
  \begin{tabular}{lcccccccc}
    \toprule
    \multirow{2}{*}{Method} & \multicolumn{4}{c}{Low-Level}& \multicolumn{4}{c}{High-Level}\\
    \cmidrule(lr){2-5} \cmidrule(lr){6-9}
     &PixCorr$\uparrow$&SSIM$\uparrow$&Alex(2)$\uparrow$&Alex(5)$\uparrow$&Incep$\uparrow$&CLIP$\uparrow$&Eff$\downarrow$&SwAV$\downarrow$\\
     \midrule

    Takagi...~\cite{takagi2023high} \scriptsize\textcolor{gray}{[CVPR'23]}
        & 0.246 & {0.410} & {78.9\%} & {85.6\%} & 83.8\% & 82.1\% & {0.811} & {0.504} \\

    MindEye1~\cite{scotti2023reconstructing} \scriptsize\textcolor{gray}{[NeurIPS'24]}
        & 0.319 & {0.360} & 92.8\% & 96.9\% & 94.6\% & 93.3\% & 0.648 & 0.377 \\

    UMBRAE~\cite{xia2024umbrae} \scriptsize\textcolor{gray}{[ECCV'24]}
        & 0.283 & {0.341} & 95.5\% & 97.0\% & 91.7\% & 93.5\% & 0.700 & 0.393 \\

    NeuroPictor~\cite{huo2024neuropictor} \scriptsize\textcolor{gray}{[ECCV'24]}
        & 0.229 & {0.375} & 96.5\% & 98.4\% & 94.5\% & 93.3\% & 0.639 & 0.350 \\

    MindBridge~\cite{wang2024mindbridge} \scriptsize\textcolor{gray}{[CVPR'24]}
        & {0.151} & {0.263} & 87.7\% & 95.5\% & 92.4\% & \textbf{94.7\%} & 0.712 & 0.418 \\

    MindEye2~\cite{mindeye2} \scriptsize\textcolor{gray}{[ICML'24]}
        & 0.322 & 0.431 & 96.1\% & \underline{98.6\%} & 95.4\% & 93.0\% & \underline{0.619} & \underline{0.344} \\

    MindTuner~\cite{gong2025mindtuner} \scriptsize\textcolor{gray}{[AAAI'25]}
        & 0.322 & 0.421 & 95.8\% & \textbf{98.8\%} & 95.6\% & \underline{93.8\%} & \textbf{0.612} & \textbf{0.340} \\

    BrainIT~\cite{brainit} \scriptsize\textcolor{gray}{[ICLR'26]}
        & 0.331 & \textbf{0.473} & \textbf{97.1\%} & \underline{98.6\%} & 94.4\% & 93.0\% & 0.648 & 0.370 \\

    \textsc{Neurons}~\cite{wang2025neurons} \scriptsize\textcolor{gray}{[ICCV'25]}
        & N/A & N/A & N/A & N/A & N/A & N/A & N/A & N/A \\

    \rowcolor{cyan!10}
    \textbf{\methodname{}}
        & \textbf{0.355} & \underline{0.457} & \underline{96.8\%} & \textbf{98.8\%} & \textbf{96.2\%} & \underline{93.8\%} & \textbf{0.612} & 0.365 \\
    \midrule

    \rowcolor{cyan!10}
    Subject 1
        & 0.424 & 0.478 & 98.0\% & 99.3\% & 96.7\% & 94.1\% & 0.606 & 0.364 \\
    \rowcolor{cyan!10}
    Subject 2
        & 0.363 & 0.455 & 97.6\% & 98.9\% & 96.2\% & 93.5\% & 0.612 & 0.360 \\
    \rowcolor{cyan!10}
    Subject 5
        & 0.321 & 0.449 & 96.1\% & 98.8\% & 96.7\% & 94.7\% & 0.602 & 0.357 \\
    \rowcolor{cyan!10}
    Subject 7
        & 0.311 & 0.446 & 95.4\% & 98.0\% & 95.3\% & 92.9\% & 0.625 & 0.378 \\
     \bottomrule
  \end{tabular}}
  \label{tab:image_nsd_full}
\end{table*}

\begin{table}[!t]
\centering
\caption{
Paired significance tests of \methodname{} against the previous SOTA method SemVideo on the CC2017 dataset, evaluated on identical test samples.
For each metric, we report the per-metric mean of both models, the paired difference $\Delta$
with its 95\% bootstrap confidence interval, and three p-values obtained from a permutation
test, a paired bootstrap, and a Wilcoxon signed-rank test ($10^4$ resamples for the
permutation and bootstrap tests).
Significance stars follow the permutation p-value:
$^{\ast}p{<}0.05$, $^{\ast\ast}p{<}0.01$, $^{\ast\ast\ast}p{<}0.001$, \textsc{ns} otherwise.
}
\resizebox{\linewidth}{!}{
  \begin{tabular}{lccccc}
    \toprule
    Metric & SemVideo~\cite{yang2026semvideo} & \methodname{}  & 95\% CI
      & $p_{\text{perm}}$ / $p_{\text{boot}}$ / $p_{\text{wil}}$ \\
    \midrule
    2-way$\uparrow$    & 0.865 & 0.871  & $[+0.0045,\,+0.0181]$
      & $0.0013^{\ast\ast}$ / $0.0002$ / $0.0005$ \\
    50-way$\uparrow$   & 0.264 & 0.279  & $[+0.0041,\,+0.0315]$
      & $0.0102^{\ast}$ / $0.0120$ / $0.0037$ \\
    \bottomrule
  \end{tabular}}
\label{tab:sigtest}
\end{table}

\section{Functional ROI Parcel Assignments}
\label{app:roi_definitions}

To make the ROI grouping of the ablation analysis in Section~\ref{subsec:roi_ablation} fully reproducible, Table~\ref{tab:roi_definitions} specifies the complete set of HCP-MMP1 parcels assigned to each of the eight functional ROIs under the Glasser multi-modal parcellation~\cite{glasser2016multi}.
Of the $V = 13{,}447$ activated voxels of subject~1, $6{,}164$ fall within these eight functional ROIs and are grouped accordingly; the remaining activated voxels lie outside the eight parcels and are not assigned to any ROI. 
Parcel label IDs are taken from the publicly available \texttt{.dlabel.nii} atlas projected onto the fsLR-32k surface. 
The first six ROIs (\textsc{face} through \textsc{semantic}) are category- or stage-selective and mutually disjoint, while \textsc{ventral} and \textsc{dorsal} denote broader visual streams and overlap with the finer parcels by construction. 
Each ROI is nonetheless masked independently in the ablation.

\begin{table}[t]
\centering
\caption{
HCP-MMP1 parcel assignments for the eight functional ROIs used in the ROI ablation analysis.
The \textsc{semantic} ROI corresponds to anterior ventral temporal cortex.
}
\label{tab:roi_definitions}
\begin{tabular}{@{}ll@{}}
\toprule
ROI & HCP-MMP1 parcels \\
\midrule
\textsc{face}         & FFC, PeEc \\
\textsc{place}        & PHA1, PHA2, PHA3, VMV1, VMV2, VMV3 \\
\textsc{motion}       & MT, MST, FST \\
\textsc{early\_visual} & V1, V2, V3, V4 \\
\textsc{object}       & LO1, LO2, LO3 \\
\textsc{semantic}     & TE1a/p, TE2a/p \\
\textsc{ventral}      & V4, V8, FFC, PIT, VVC, VMV1 \\
\textsc{dorsal}       & V3A, V6, V6A, V7 \\
\bottomrule
\end{tabular}
\end{table}

\section{Additional Details for Neural-to-Visual Interpretability}
\label{app:neuro_supp}

This section provides the mathematical and implementation details that support
the neural-to-visual interpretability analyses in Sec.~\ref{sec:neuro_analysis}
of the main paper. The main text retains the key figures, quantitative effects,
and conclusions, while the derivations and methodological caveats are collected
here to keep the result narrative concise.

\subsection{Delta-fMRI Perturbation Normalization}
\label{app:delta_fmri_details}

The delta-fMRI analysis in Sec.~\ref{subsec:delta_fmri} is designed as a
qualitative direction-consistency probe rather than a metrically calibrated
test of proportionality. The reason is that each perturbed response is
z-score normalized independently before being passed to the encoder.

Before normalization, the directional perturbation is
\begin{equation}
    \tilde{\mathbf{x}}_j(\alpha)
    =
    \mathbf{x}_j+\alpha\mathbf{d},
    \quad
    \mathbf{d}=\mathbf{x}_j-\mathbf{x}_i ,
\end{equation}
so all raw perturbations lie on a single line and are ordered by $\alpha$.
Negative $\alpha$ values interpolate from $\mathbf{x}_j$ toward
$\mathbf{x}_i$, whereas positive $\alpha$ values extrapolate beyond
$\mathbf{x}_j$ along the same direction.

The per-response z-score normalization is
\begin{equation}
    \mathcal{Z}(\mathbf{v})
    =
    \frac{\mathbf{v}-\mu(\mathbf{v})\mathbf{1}}
    {\sigma(\mathbf{v})},
\end{equation}
where $\mu(\mathbf{v})$ and $\sigma(\mathbf{v})$ denote the voxel-wise mean
and standard deviation of $\mathbf{v}$. Writing
$\mu(\tilde{\mathbf{x}}_j(\alpha))=\mu_j+\alpha\mu_d$ and
$\sigma(\tilde{\mathbf{x}}_j(\alpha))=\sigma(\alpha)$ gives
\begin{equation}
\mathcal{Z}(\tilde{\mathbf{x}}_j(\alpha))
=
\frac{1}{\sigma(\alpha)}
\Big[
(\mathbf{x}_j-\mu_j\mathbf{1})
+
\alpha(\mathbf{d}-\mu_d\mathbf{1})
\Big].
\end{equation}
The bracketed term moves along a fixed mean-centered direction, while the
prefactor $1/\sigma(\alpha)$ introduces an $\alpha$-dependent rescaling.
Therefore, strict collinearity is not exactly preserved after normalization.

For the small perturbations used in the main text ($|\alpha|\le0.5$), a
first-order approximation gives
\begin{equation}
\label{eq:delta_zscore_firstorder_supp}
\mathcal{Z}(\tilde{\mathbf{x}}_j(\alpha))
\approx
\mathcal{Z}(\mathbf{x}_j)
+
\frac{\alpha}{\sigma_j}
(\mathbf{d}-\mu_d\mathbf{1}).
\end{equation}
To this order, the normalized response moves along a single fixed direction,
and the sign of the displacement follows the sign of $\alpha$. Thus,
interpolation and extrapolation remain directionally meaningful even though
the path is not metrically uniform.

We also explicitly bound the residual rescaling introduced by the
$\alpha$-dependent standard deviation. Expanding the per-response variance
gives
\begin{equation}
    \sigma(\alpha)^2
    =
    \sigma_j^2
    +
    2\alpha\,\mathrm{cov}_k(x_{j,k},d_k)
    +
    \alpha^2\sigma_d^2 ,
\end{equation}
where $\mathrm{cov}_k$ and $\sigma_d^2$ are the voxel-wise covariance and
variance of $\mathbf{x}_j$ and $\mathbf{d}$. Hence the relative deviation of
the normalization factor is
$\left|\sigma(\alpha)/\sigma_j-1\right|
=\mathcal{O}(|\alpha|\sigma_d/\sigma_j)$ to first order.
Empirically, over all clip pairs $(i,j)$ and all perturbation strengths used
in the analysis,
\begin{equation}
    \max_{i,j,\alpha}
    \left|
    \frac{\sigma(\alpha)}{\sigma_j}-1
    \right|
    =
    0.024 .
\end{equation}
This small deviation supports treating the perturbation path as directionally
meaningful. However, because equal increments in $\alpha$ do not necessarily
correspond to equal Euclidean steps after per-response normalization, we do
not interpret the visual changes as a quantitative proportionality curve. A
fully calibrated analysis would require a metric-preserving perturbation
scheme, for example applying fixed training-set normalization statistics
rather than per-response normalization.

\subsection{ROI Ablation Read-Out and Statistical Controls}
\label{app:roi_ablation_details}

The ROI-ablation analysis in Sec.~\ref{subsec:roi_ablation} probes the
sensitivity of the trained fMRI encoder read-out to functional cortical
regions. The read-out is
\begin{equation}
    f
    =
    \mathcal{P}_i
    \circ
    \mathcal{B}_{\theta}
    \circ
    \mathrm{ridge},
\end{equation}
where $\mathrm{ridge}$ is the voxel-to-latent ridge projection,
$\mathcal{B}_{\theta}$ is the residual-MLP backbone, and $\mathcal{P}_i$ is
the image-level projection head that produces the fMRI-derived CLIP visual
embedding. The downstream temporal projection $\mathcal{P}_{vid}$, text head
$\mathcal{P}_t$, diffusion prior $\mathcal{M}$, and final diffusion generator
are not included in $f$. This isolates the neural encoder read-out from the
additional stochasticity and generative-prior effects of the final image/video
synthesis pipeline.

Accordingly, the ROI sensitivity score
\begin{equation}
    \Delta \cos_r(\mathbf{x}_i)
    =
    1 -
    \cos\big(
    f(\mathbf{x}_i),
    f(\mathbf{x}_i\odot\mathbf{m}_r)
    \big)
\end{equation}
measures how much ablating ROI $r$ changes the predicted CLIP visual
embedding, not how much the final generated image or video changes. A larger
$\Delta\cos_r$ indicates that ROI $r$ carries information used by the
encoder, but it does not directly quantify reconstruction degradation after
the downstream diffusion generator. We choose the embedding-level score
because it provides a cleaner measurement of neural read-out dependence.
End-to-end ROI ablation through the full stochastic generation pipeline is a
natural extension for future work.

The read-out $f$ should also not be interpreted as a frozen or generic
CLIP-alignment encoder. In \methodname{}, the ridge map, backbone
$\mathcal{B}_{\theta}$, image-level projection head $\mathcal{P}_i$, and
temporal projection $\mathcal{P}_{vid}$ are trained as part of the full model.
The Serial Visual Decoupler operates on $\mathbf{e}^{v}_c$ without gradient
detachment, so the decoupling losses backpropagate through the encoder.
Therefore, the ROI-ablation analysis probes the read-out of the fully trained
\methodname{} encoder, jointly shaped by both the embedding-alignment
objective $\mathcal{L}_{enc}$ and the cortex-inspired decoupling objective
$\mathcal{L}_{dec}$.

\paragraph{Matched, mismatched, and permutation-null correlations.}
For each test sample $\mathbf{x}_i$, the main analysis computes the matched
Spearman correlation
\begin{equation}
    \rho_{\mathrm{m}}(i)
    =
    \mathrm{Spearman}
    \big(
    E_{\cdot}(\mathbf{x}_i),
    \Delta\cos_{\cdot}(\mathbf{x}_i)
    \big),
\end{equation}
where $E_{\cdot}(\mathbf{x}_i)$ is the eight-ROI activation-energy profile and
$\Delta\cos_{\cdot}(\mathbf{x}_i)$ is the corresponding eight-ROI ablation
profile. This statistic asks whether the ROIs most activated by the stimulus
are also the ROIs to which the encoder read-out is most sensitive.

To assess stimulus specificity, we compare the matched statistic against a
paired permutation null. Each of the $1{,}000$ permutations $\pi$ is a
bijection over the $n$ test samples, and the null statistic is
\begin{equation}
    \frac{1}{n}
    \sum_i
    \mathrm{Spearman}
    \big(
    E_{\cdot}(\mathbf{x}_i),
    \Delta\cos_{\cdot}(\mathbf{x}_{\pi(i)})
    \big).
\end{equation}
This preserves the marginal distribution of both energy profiles and ablation
profiles while destroying their sample-wise correspondence. The empirical
$p$-value reported in the main text is computed against this permutation null.

For visualization, we also report a descriptive mismatched baseline that pairs
energy and ablation profiles from different samples,
\begin{equation}
    \rho_{\mathrm{mm}}(i,j)
    =
    \mathrm{Spearman}
    \big(
    E_{\cdot}(\mathbf{x}_i),
    \Delta\cos_{\cdot}(\mathbf{x}_j)
    \big),
    \quad i\ne j .
\end{equation}
The matched-minus-null gap and the matched-versus-mismatched effect size thus
serve slightly different purposes: the former is used for statistical
significance, while the latter provides an interpretable descriptive effect
size.

\paragraph{Random-grouping control.}
To rule out the possibility that the ablation effect is driven by ROI size or
generic contiguous voxel groups, we repeat the analysis using $1{,}000$
size-matched random partitions. The $6{,}164$ ROI-assigned voxels are randomly
permuted and divided into eight groups with the same sizes as the real Glasser
ROIs. For each random partition, we recompute the matched-minus-mismatched
specificity gap. The real Glasser parcellation produces a gap of $0.364$,
whereas the random-grouping null is $0.061\pm0.010$, and no random partition
reaches the real gap. This confirms that the observed specificity depends on
functional ROI structure rather than ROI size alone.

\paragraph{Voxel-level attribution.}
The voxel-level attribution map in Sec.~\ref{subsec:roi_ablation} is computed
from the positive input-gradient product. Let
\begin{equation}
    Y(\mathbf{x})=\lVert f(\mathbf{x})\rVert^2
\end{equation}
denote the squared norm of the predicted CLIP visual embedding. We define
\begin{equation}
    \mathbf{a}(\mathbf{x})
    =
    \mathrm{ReLU}
    \big(
    \mathbf{x}\odot\nabla_{\mathbf{x}}Y(\mathbf{x})
    \big).
\end{equation}
This attribution retains voxels whose positive input-gradient product
contributes to the model response. Projecting $\mathbf{a}(\mathbf{x})$ onto
the cortical surface provides a voxel-level visualization that complements
the ROI-level ablation statistics.

\subsection{Details and Limitations of ROI Amplification}
\label{app:roi_boost_details}

The ROI amplification analysis in Sec.~\ref{subsec:roi_enhance} tests whether
selectively enhancing a functional ROI can steer the decoded visual semantics
toward the category associated with that ROI. Given a target ROI
$\mathcal{R}$ and an original trial-averaged fMRI response
$\mathbf{x}\in\mathbb{R}^V$, we construct a manipulated response by setting
all voxels in $\mathcal{R}$ to a fixed high activation value $v=3.0$ and
suppressing non-target voxels by a factor of $0.1$:
\begin{equation}
    \tilde{x}_k =
    \begin{cases}
    v, & k\in\mathcal{R},\\
    0.1x_k, & k\notin\mathcal{R}.
    \end{cases}
\end{equation}
The manipulated response is then z-score normalized and decoded through the
same encoder and generative pipeline as the original response.

This intervention is intentionally strong. It bundles three operations:
target-ROI saturation, non-target suppression, and global z-score
renormalization. Therefore, the result should be interpreted as a
region-selective, directionally specific semantic modulation test, not as a
calibrated estimate of the causal gain of the target ROI alone. In other
words, the experiment asks whether pushing the neural input into a
face-dominant or place-dominant direction changes the decoded content in the
expected semantic direction; it does not estimate how much natural variation
in a single ROI would change perception or reconstruction.

This design also makes the intervention partially out-of-distribution relative
to the natural fMRI responses used for training. The instability observed for
one ROI direction in each of the two additional CC2017 subjects
(Table~\ref{tab:app_enhance}) should therefore be interpreted cautiously. The
same subjects still show strong ROI-ablation stimulus specificity
(Table~\ref{tab:app_ablation}), indicating that the encoder reads
stimulus-relevant ROI information. The sign instability is specific to the
strong boost intervention and its propagation through the downstream decoder.
A milder in-distribution intervention, such as scaling only the target ROI by
a modest factor without suppressing the remaining voxels, would provide a
cleaner test of graded ROI gain and is left to future work.

\section{Zero-Shot CLIP Scoring of Face and Place Content}
\label{app:clip_scoring}

To score the face and place content of each reconstructed keyframe in Section~\ref{subsec:roi_enhance}, we use a zero-shot CLIP classifier.

For each target category $c \in \{\textsc{face}, \textsc{place}\}$, we specify a set of positive prompts $\mathcal{P}_c^{+}$ describing the category and a set of negative prompts $\mathcal{P}_c^{-}$ describing competing content (Table~\ref{tab:clip_prompts}). 
Each prompt $p$ and each reconstructed keyframe are embedded with the CLIP text and image encoders and $\ell_2$-normalized, yielding unit vectors $\mathbf{t}_p$ and $\mathbf{z}$. 
We compute the cosine similarity between the image and each prompt, and average the similarities within the positive and negative sets:
\begin{equation}
  \bar{s}_c^{\pm}(\mathbf{z}) =
  \frac{1}{|\mathcal{P}_c^{\pm}|}
  \sum_{p \in \mathcal{P}_c^{\pm}} \cos(\mathbf{z}, \mathbf{t}_p).
  \label{eq:clip_simmean}
\end{equation}
The two pooled similarities are converted into a category score by a two-way softmax with a fixed inverse temperature $\beta = 100$:
\begin{equation}
  \begin{aligned}
  s_c(\mathbf{z}) &=
  \sigma\!\big(\beta[\,\bar{s}_c^{+}(\mathbf{z}) - \bar{s}_c^{-}(\mathbf{z})\,]\big) \\
  &=
  \frac{\exp\!\big(\beta\,\bar{s}_c^{+}(\mathbf{z})\big)}
       {\exp\!\big(\beta\,\bar{s}_c^{+}(\mathbf{z})\big)
        + \exp\!\big(\beta\,\bar{s}_c^{-}(\mathbf{z})\big)},
  \end{aligned}
  \label{eq:clip_zeroshot}
\end{equation}
where $\sigma(\cdot)$ is the logistic sigmoid. 
The per-clip score is obtained by averaging $s_c$ over the keyframes of the clip.

\begin{table}[t]
\centering
\caption{
Zero-shot CLIP text prompts for scoring face and place content.
Each score contrasts the mean similarity of the positive prompts against that of the negative prompts.
}
\label{tab:clip_prompts}
\begin{tabular}{@{}ll@{}}
\toprule
\multicolumn{2}{@{}l}{\textbf{\textsc{Face} score}} \\
\midrule
Positive & ``a photo of a human face'' \\
         & ``a close-up portrait of a person'' \\
         & ``a person looking at the camera'' \\
Negative & ``a landscape without people'' \\
         & ``an empty outdoor scene'' \\
         & ``a photo of buildings and architecture'' \\
\midrule
\multicolumn{2}{@{}l}{\textbf{\textsc{Place} score}} \\
\midrule
Positive & ``a photo of a building'' \\
         & ``an indoor room or hallway'' \\
         & ``a scenic landscape with architecture'' \\
Negative & ``a close-up of a face'' \\
         & ``a single isolated object'' \\
         & ``an abstract pattern'' \\
\bottomrule
\end{tabular}
\end{table}

\section{Additional Ablation: GRPO-based Image-to-Image Refinement}
\label{subsec:grpo_refinement}

\begin{figure}[!t]
    \centering
    \includegraphics[width=\linewidth]{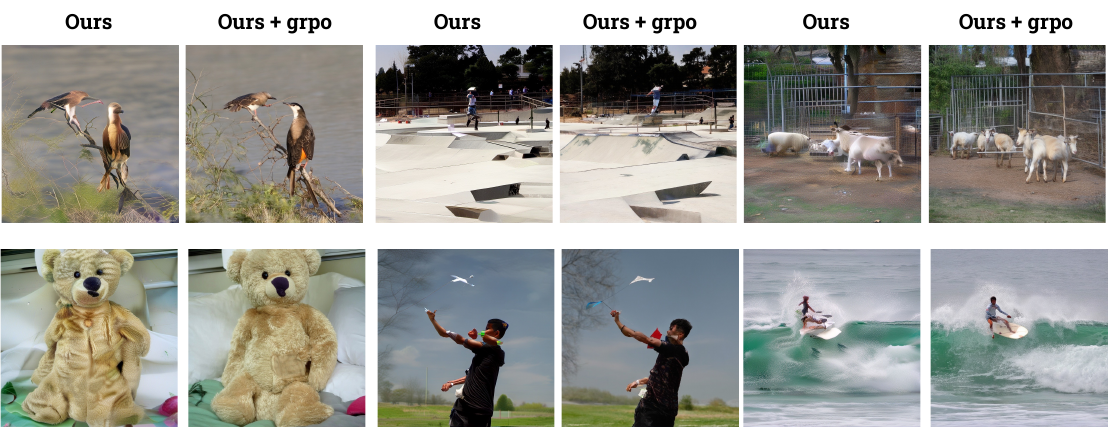}
    \caption{Qualitative examples of GRPO-based image-to-image refinement. The refinement often improves semantic clarity and reduces visual ambiguity, but does not consistently improve automatic reconstruction metrics.}
    \label{fig:grpo_refinement}
\end{figure}

We further explore whether the reconstructed images can be improved through a post-hoc image-to-image refinement stage. Specifically, we fine-tune SD3.5 with GRPO and use the resulting model to refine our reconstructed images, as shown in Fig.~\ref{fig:grpo_refinement}. Qualitatively, this refinement often produces sharper and more semantically faithful images. In many cases, it reduces visual ambiguity and recovers clearer object-level details compared with the original reconstruction.

However, this improvement is not consistently reflected by automatic metrics. Although the refined images are visually more interpretable in several examples, the overall quantitative scores do not improve. We hypothesize that the GRPO-refined model tends to enhance semantic plausibility and perceptual clarity, while also introducing distributional changes or hallucinated details that are not always aligned with the ground-truth stimulus under pixel-level or embedding-based evaluation metrics. Therefore, we use this experiment as an additional qualitative analysis rather than the main quantitative result.

\section{Additional Subjects and Datasets for ROI Analyses}
\label{app:additional_subjects}

The ROI ablation (Section~\ref{subsec:roi_ablation}) and ROI amplification
(Section~\ref{subsec:roi_enhance}) analyses in the main text are reported on
a single subject for clarity of exposition. To establish that the reported
effects are not idiosyncratic to one subject or one acquisition protocol, we
replicate both analyses on additional CC2017 subjects and on all three
subjects of the HCP movie-watching dataset. The HCP data differ from CC2017
in scanner, preprocessing, stimulus set, and the number of activated
grayordinates, and therefore provide an independent test of the same
hypotheses. Across all subjects and datasets we use the identical pipeline,
ROI definitions, and statistical tests as in the main text; only the
per-subject brain encoder and the per-subject activated-voxel set change.

\subsection{ROI Ablation Generalizes Across Subjects and Datasets}
\label{app:ablation_additional}

For each subject we repeat the per-sample stimulus-specificity test of
Section~\ref{subsec:roi_ablation}: for every stimulus we correlate the
encoder's eight-ROI ablation profile $\Delta\cos_{\cdot}(\mathbf{x}_i)$ with
the corresponding fMRI activation-energy profile $E_{\cdot}(\mathbf{x}_i)$
(matched Spearman $\rho_{\rm m}$), and assess significance with the same
$1{,}000$-permutation, bijective paired-permutation null. A positive
matched-minus-null gap $\Delta\rho$ indicates that the encoder relies most on
the ROIs that the stimulus most strongly drives.

As summarized in Table~\ref{tab:app_ablation}, the effect is large, positive,
and highly significant for \emph{every} subject in \emph{both} datasets: the
matched-minus-null gap ranges from $\Delta\rho=0.33$ to $0.46$ with a pooled-SD
Cohen's $d$ (matched vs.\ mismatched distribution) between $0.86$ and $1.27$,
and the empirical $p$-value reaches the
$1/1001$ permutation floor ($p<10^{-3}$) in all cases. The consistency of
this result across two datasets with different acquisition protocols, voxel
counts, and stimulus distributions indicates that stimulus-specific ROI
dependence is a robust and reproducible property of the learned encoder
rather than an artifact of any single subject.

We note that the \emph{population-level} ROI ranking (the single Spearman
correlation between the across-sample mean ablation profile and mean energy
profile, last column of Table~\ref{tab:app_ablation}) is more variable and is
in some HCP subjects negative. This population statistic is dominated by a
few high-energy early-visual ROIs and aggregates away the per-stimulus
structure; it is therefore a coarse summary and is not the quantity our
specificity test targets. The per-sample matched-versus-null gap, which
preserves the stimulus-by-stimulus correspondence, is the appropriate measure
of stimulus specificity and is uniformly strong.

\begin{table}[t]
\centering
\caption{ROI-ablation stimulus specificity across subjects and datasets.
  $V$: number of activated voxels. $\bar{\rho}_{\rm m}$: mean matched
  per-sample Spearman correlation. Null: mean of the paired-permutation null.
  $\Delta\rho$: matched-minus-null gap. $d$: pooled-SD Cohen's $d$ between the
  matched and mismatched per-sample $\rho$ distributions. All
  empirical $p<10^{-3}$ ($1{,}000$ permutations). The population column is the
  coarse across-sample ROI ranking correlation (see text).}
\label{tab:app_ablation}
\resizebox{\columnwidth}{!}{%
\begin{tabular}{llccccc}
\toprule
Dataset & Subj & $V$ & $\bar{\rho}_{\rm m}$ & $\Delta\rho$ & $d$ & pop.\ $\rho$ \\
\midrule
CC2017 & S1$^{\dagger}$ & $13{,}447$ & $0.631$ & $0.368$ & $1.11$ & $+0.50$ \\
CC2017 & S2 & $14{,}828$ & $0.573$ & $0.393$ & $1.11$ & $+0.55$ \\
CC2017 & S3 & $\phantom{0}9{,}114$ & $0.519$ & $0.434$ & $1.22$ & $+0.26$ \\
\midrule
HCP    & S1 & $12{,}000$ & $0.337$ & $0.325$ & $0.86$ & $-0.76$ \\
HCP    & S2 & $12{,}000$ & $0.402$ & $0.457$ & $1.27$ & $-0.81$ \\
HCP    & S3 & $13{,}292$ & $0.373$ & $0.415$ & $1.13$ & $+0.14$ \\
\bottomrule
\end{tabular}%
}
\\[2pt]
{\footnotesize $^{\dagger}$The subject reported in the main text
(Section~\ref{subsec:roi_ablation}).}
\end{table}

\subsection{ROI Amplification Across Subjects and Datasets}
\label{app:enhance_additional}

We next replicate the ROI-boost double-dissociation experiment of
Section~\ref{subsec:roi_enhance}. For each subject we boost the \textsc{face}
ROI and, separately, the \textsc{place} ROI, and score every reconstructed
keyframe with the same zero-shot CLIP classifier
(Appendix~\ref{app:clip_scoring}). For a clean double dissociation, boosting a
region should raise its matched category score (the on-diagonal entries of the
$2\times2$ score matrix) while leaving or suppressing the competing category
(the off-diagonal entries). All comparisons use the Wilcoxon signed-rank test
against the unmanipulated reconstruction.

\paragraph{HCP subjects show a clean, reproducible double dissociation.}
On all three HCP subjects the same asymmetric double dissociation reported in
the main text reproduces cleanly (top block of Table~\ref{tab:app_enhance}):
boosting the face ROI raises CLIP-face (on-diagonal $d=+0.88$ to $+1.07$)
while lowering CLIP-place, and boosting the place ROI raises CLIP-place
(on-diagonal $d=+1.03$ to $+1.35$) while lowering CLIP-face. Both on-diagonal
effects are positive and highly significant for every HCP subject, and the
off-diagonal entries are uniformly suppressive, exactly matching the
competitive, region-selective pattern described in the main text. Because the
HCP data were acquired and preprocessed independently of CC2017, this
constitutes a strong cross-dataset confirmation that targeted ROI
amplification steers the reconstruction toward the matched semantic category.

\paragraph{Additional CC2017 subjects.}
We also apply the boost intervention to two further CC2017 subjects
(bottom block of Table~\ref{tab:app_enhance}). In each case one of the two ROI
directions reproduces the intended on-diagonal lift with a large effect
(CC2017~S2 place: $d=+1.57$; CC2017~S3 face: $d=+0.80$), confirming that the
boost can drive the matched category. The complementary direction in each
subject is, however, slightly imperfect: the on-diagonal effect is mildly
negative (CC2017~S2 face: $d=-0.77$; CC2017~S3 place: $d=-0.94$), so boosting
the region does not raise its own matched category as cleanly. We note that
these same two subjects show among the strongest ablation specificity in the
study (Table~\ref{tab:app_ablation}), so the encoder clearly reads the relevant
ROI signal; the amplification direction simply does not always propagate with a
consistent sign through the decoder. This is a minor instability that affects
only a single ROI direction per subject, and reducing the boost magnitude does
not remove it, so we report it for completeness rather than treating it as a
failure of the underlying representation.

\begin{table}[t]
\centering
\caption{ROI-boost double dissociation across subjects and datasets.
  Each subject contributes two rows (face boost, place boost). On-diagonal entries (the category matched to the boosted ROI) are in \textbf{bold} and are the intended effect; off-diagonal entries report the competing category.
  Cell values are $\Delta$ in CLIP score (Cohen's $d$ in parentheses),
  Wilcoxon signed-rank vs.\ the unmanipulated reconstruction. The HCP block uses $N{=}200$ samples per ROI and the CC2017 block $N{=}200$; all reported effects have $p<10^{-2}$. Reversed on-diagonal entries (boosting a region lowers its own matched category) are marked $^{\ddagger}$.}
\label{tab:app_enhance}
\resizebox{\columnwidth}{!}{%
\begin{tabular}{lllcc}
\toprule
Dataset & Subj & Boost & $\Delta$ CLIP-face & $\Delta$ CLIP-place \\
\midrule
HCP & S1 & Face  & $\mathbf{+0.255}\ (\mathbf{+0.88})$ & $-0.161\ (-0.67)$ \\
HCP & S1 & Place & $-0.424\ (-1.22)$ & $\mathbf{+0.306}\ (\mathbf{+1.03})$ \\
HCP & S2 & Face  & $\mathbf{+0.290}\ (\mathbf{+1.07})$ & $-0.264\ (-1.15)$ \\
HCP & S2 & Place & $-0.426\ (-1.26)$ & $\mathbf{+0.349}\ (\mathbf{+1.23})$ \\
HCP & S3 & Face  & $\mathbf{+0.318}\ (\mathbf{+1.07})$ & $-0.325\ (-1.21)$ \\
HCP & S3 & Place & $-0.533\ (-1.37)$ & $\mathbf{+0.447}\ (\mathbf{+1.35})$ \\
\midrule
CC2017 & S2 & Face  & $\mathbf{-0.237}\ (\mathbf{-0.77})^{\ddagger}$ & $-0.034\ (-0.17)$ \\
CC2017 & S2 & Place & $-0.125\ (-0.43)$ & $\mathbf{+0.494}\ (\mathbf{+1.57})$ \\
CC2017 & S3 & Face  & $\mathbf{+0.205}\ (\mathbf{+0.80})$ & $-0.133\ (-0.64)$ \\
CC2017 & S3 & Place & $+0.257\ (+0.98)$ & $\mathbf{-0.207}\ (\mathbf{-0.94})^{\ddagger}$ \\
\bottomrule
\end{tabular}%
}
\end{table}

\subsection{Per-Subject Figures}
\label{app:additional_figures}

\begin{figure*}[p]
  \centering
  \includegraphics[width=0.80\linewidth]{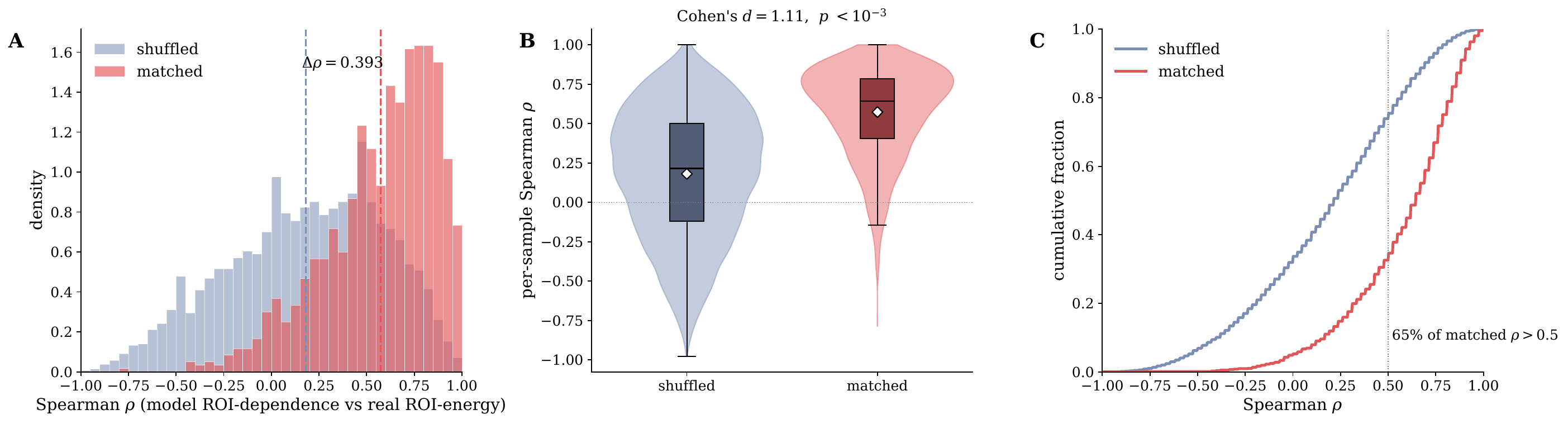}\\[2pt]
  \includegraphics[width=0.80\linewidth]{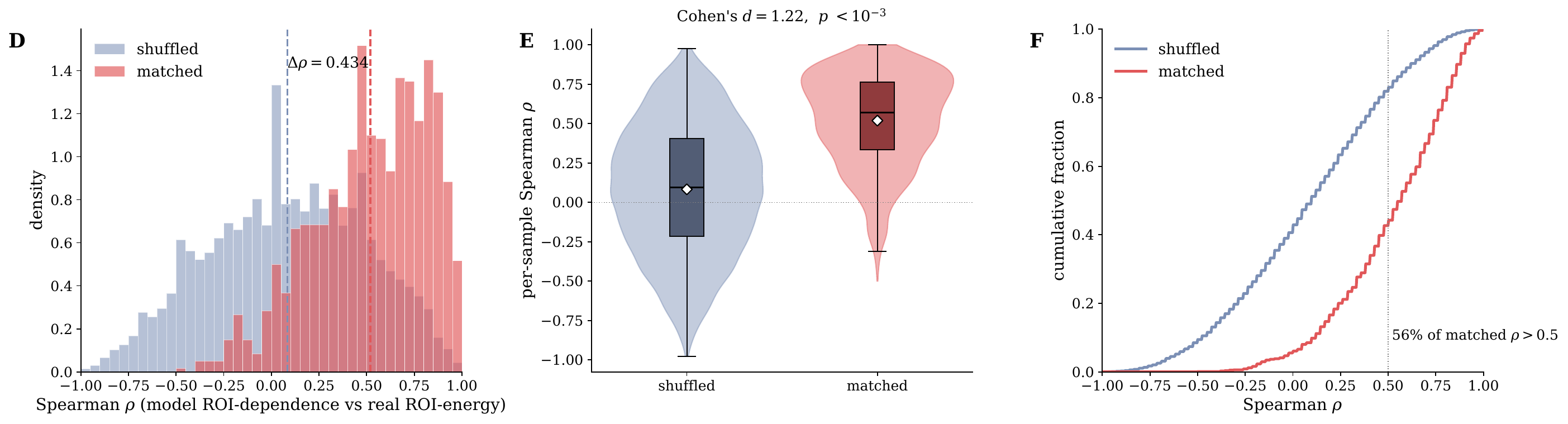}
\caption{%
  \textbf{Per-sample ROI-ablation stimulus specificity for the additional CC2017 subjects.} Panels A--C show CC2017~S2, and panels D--F show CC2017~S3. For each subject, the three panels respectively report the matched vs.\ mismatched Spearman-$\rho$ histogram, the violin/box comparison of per-sample Spearman-$\rho$, and the empirical CDF. In both subjects, the matched distribution is strongly right-shifted relative to the permutation-based null, consistent with Table~\ref{tab:app_ablation}.%
}
  \label{fig:app_spec_cc2017}
\end{figure*}

\begin{figure*}[p]
  \centering
  \includegraphics[width=0.80\linewidth]{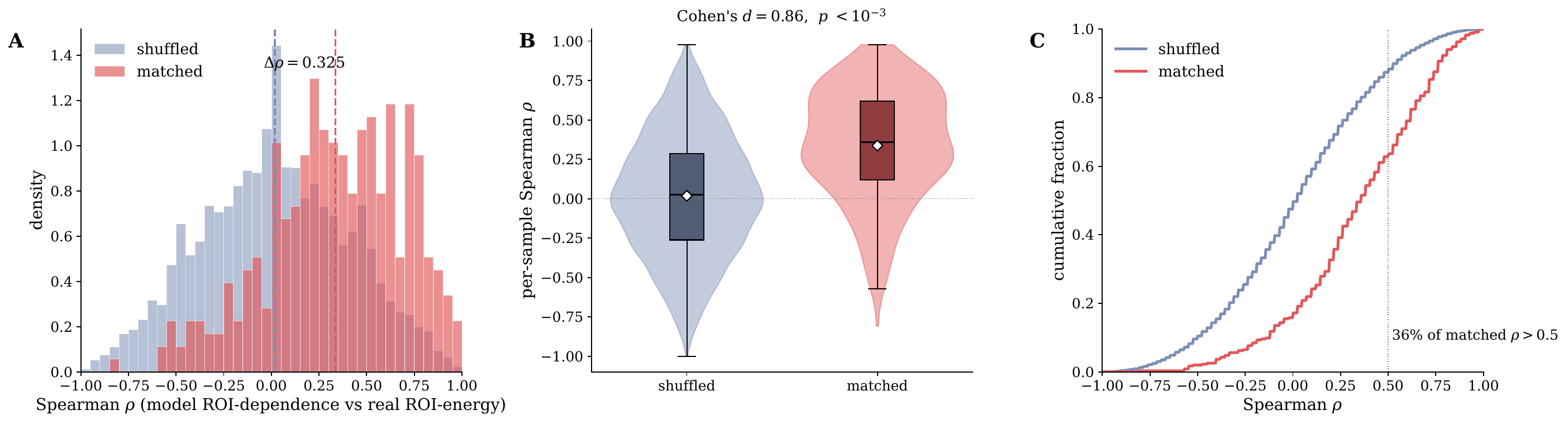}\\[2pt]
  \includegraphics[width=0.80\linewidth]{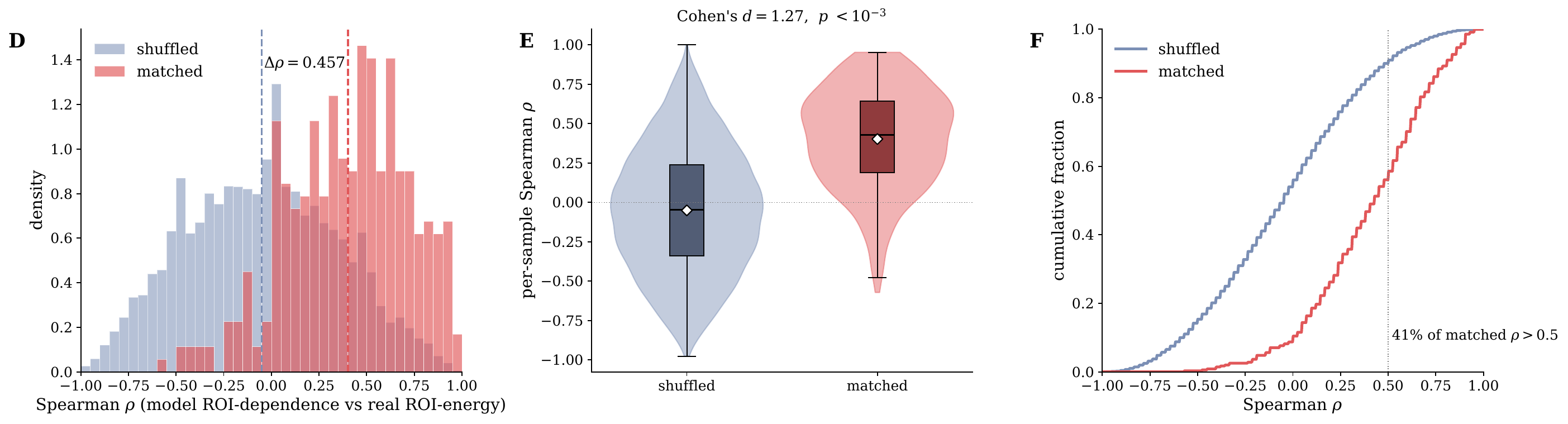}\\[2pt]
  \includegraphics[width=0.80\linewidth]{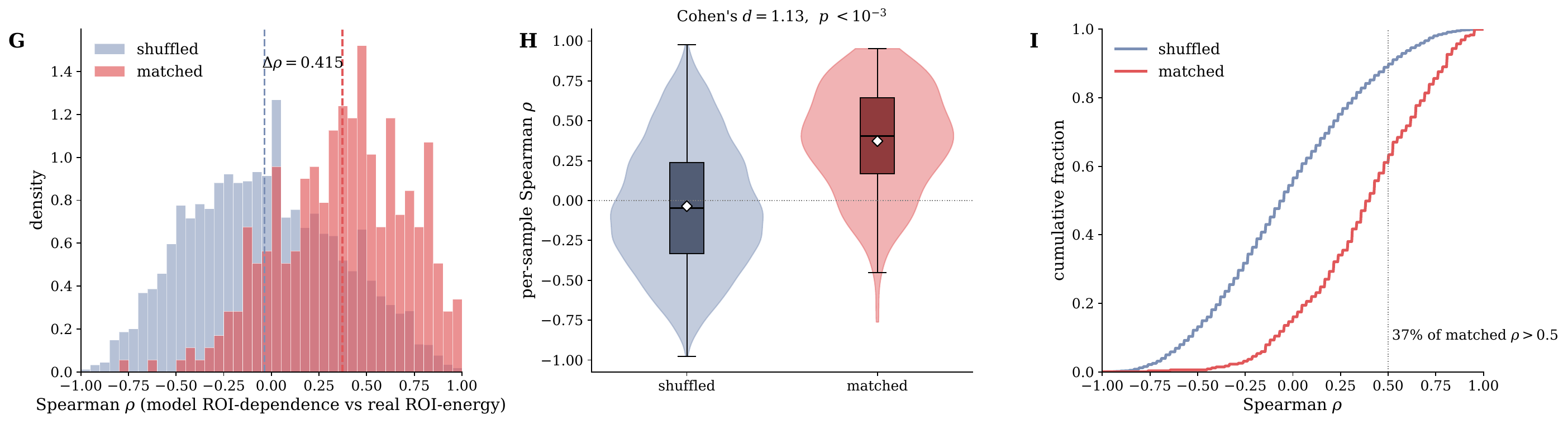}
\caption{%
  \textbf{Per-sample ROI-ablation stimulus specificity for the HCP subjects.}
  Panels A--C show HCP~S1, panels D--F show HCP~S2, and panels G--I show HCP~S3. For each subject, the three panels respectively report the matched vs.\ mismatched Spearman-$\rho$ histogram, the violin/box comparison of per-sample Spearman-$\rho$, and the empirical CDF. In all three subjects, the matched distribution is strongly right-shifted relative to the permutation-based null, consistent with Table~\ref{tab:app_ablation}.%
}
  \label{fig:app_spec_hcp}
\end{figure*}

\begin{figure}[t]
  \centering
  \includegraphics[width=\linewidth]{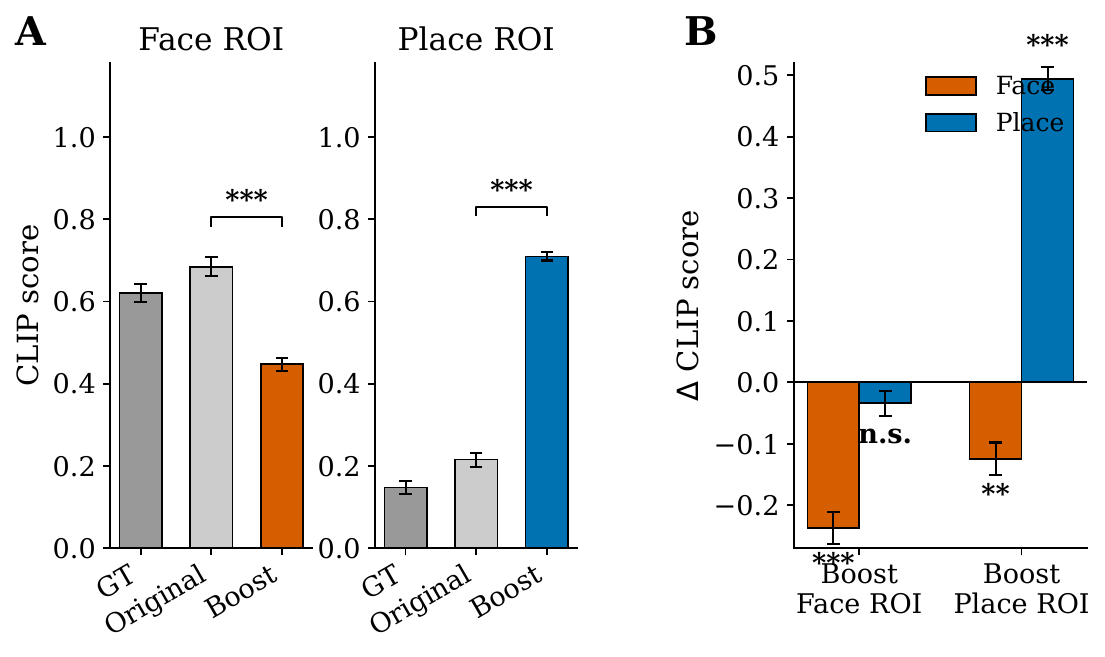}\\[4pt]
  \includegraphics[width=\linewidth]{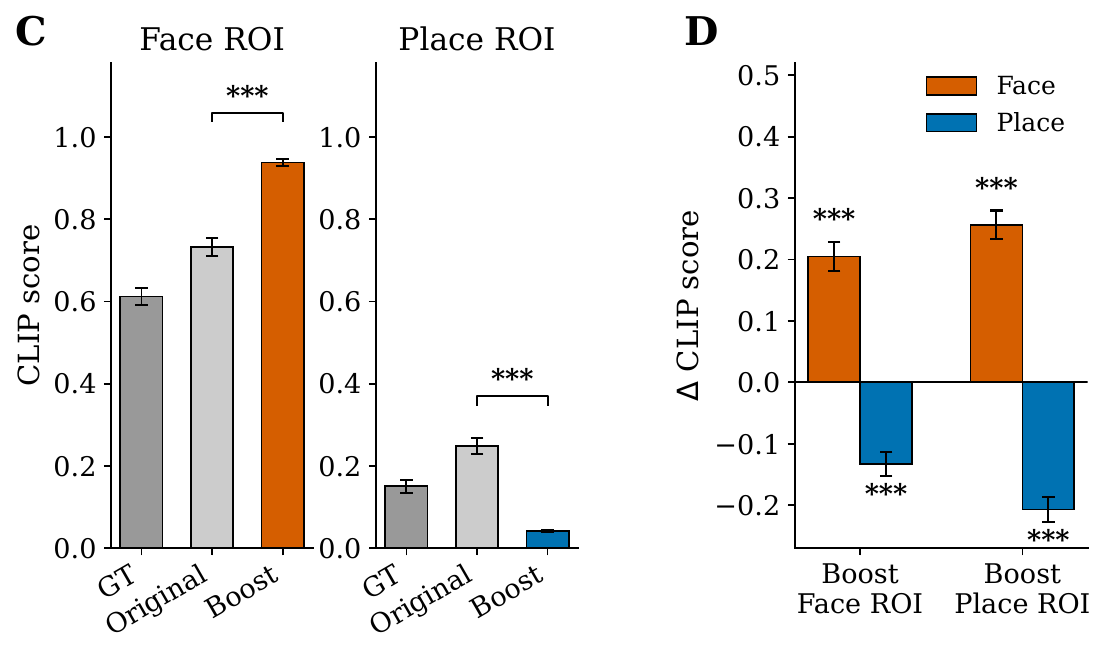}
\caption{%
  \textbf{ROI-boost double-dissociation summaries for the additional CC2017
  subjects.} Panels A--B show CC2017~S2, and panels C--D show CC2017~S3.
  For each subject, the left panel reports the matched-category lift after boosting the corresponding ROI, and the right panel reports the cross-suppressive $\Delta$-score bars. These panels reproduce the per-subject analogue of main-text Fig.~\ref{fig:boost_and_dissociation}.%
}
  \label{fig:app_dd_cc2017}
\end{figure}

\begin{figure}[t]
  \centering
  \includegraphics[width=\linewidth]{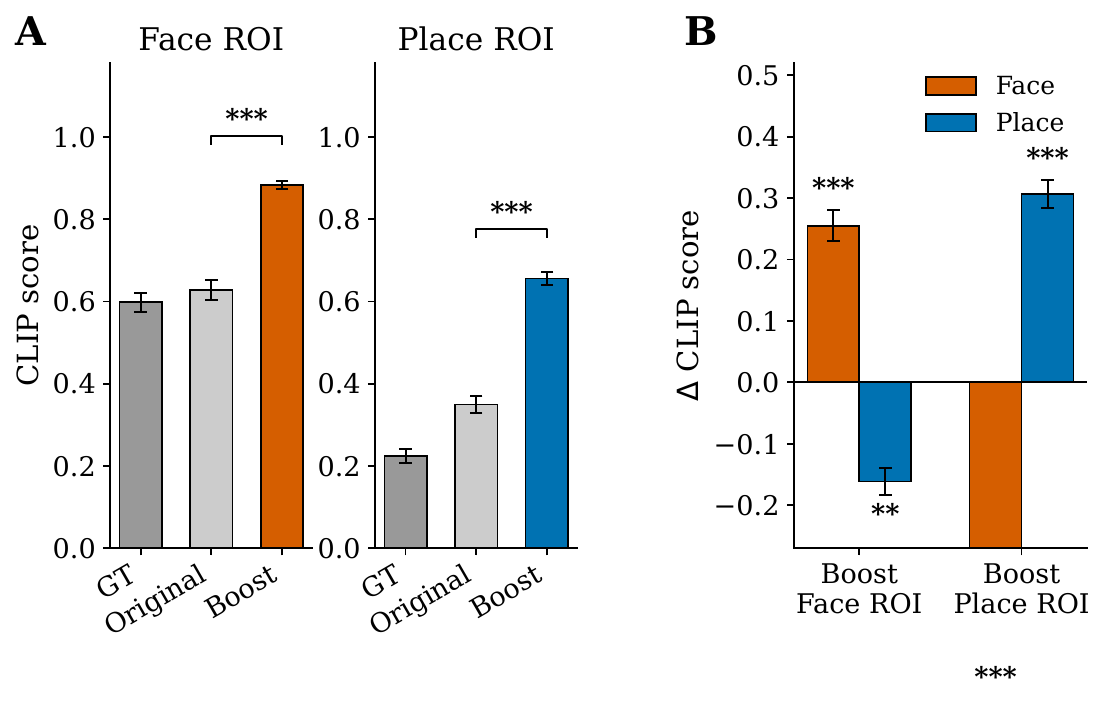}\\[4pt]
  \includegraphics[width=\linewidth]{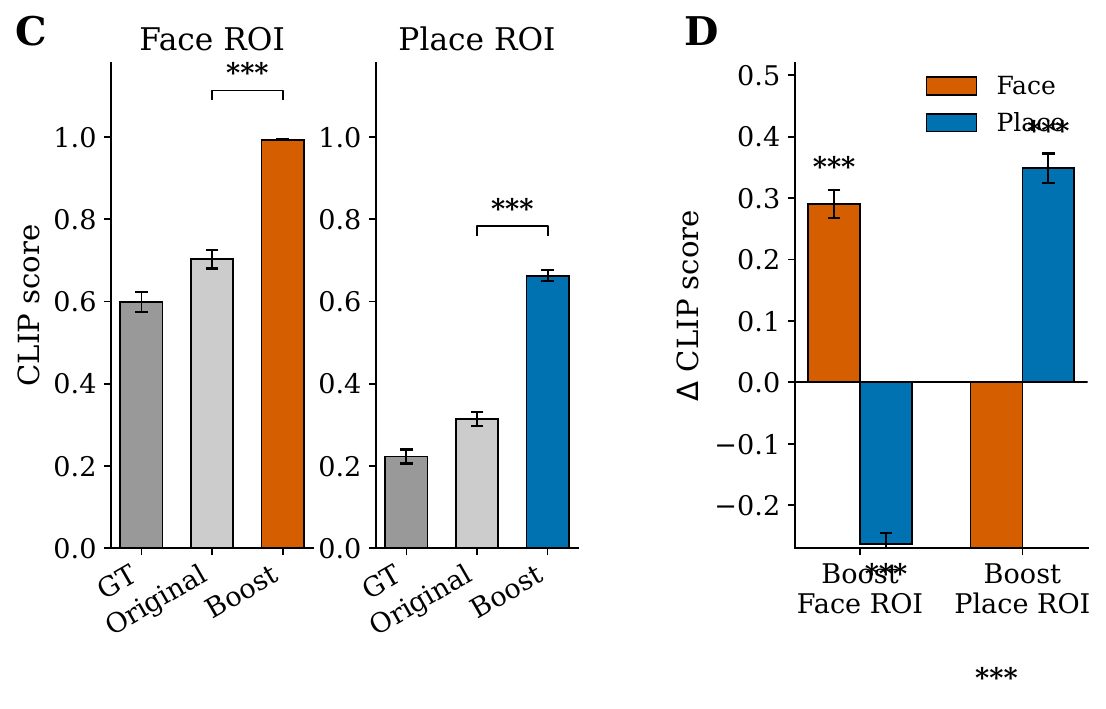}\\[4pt]
  \includegraphics[width=\linewidth]{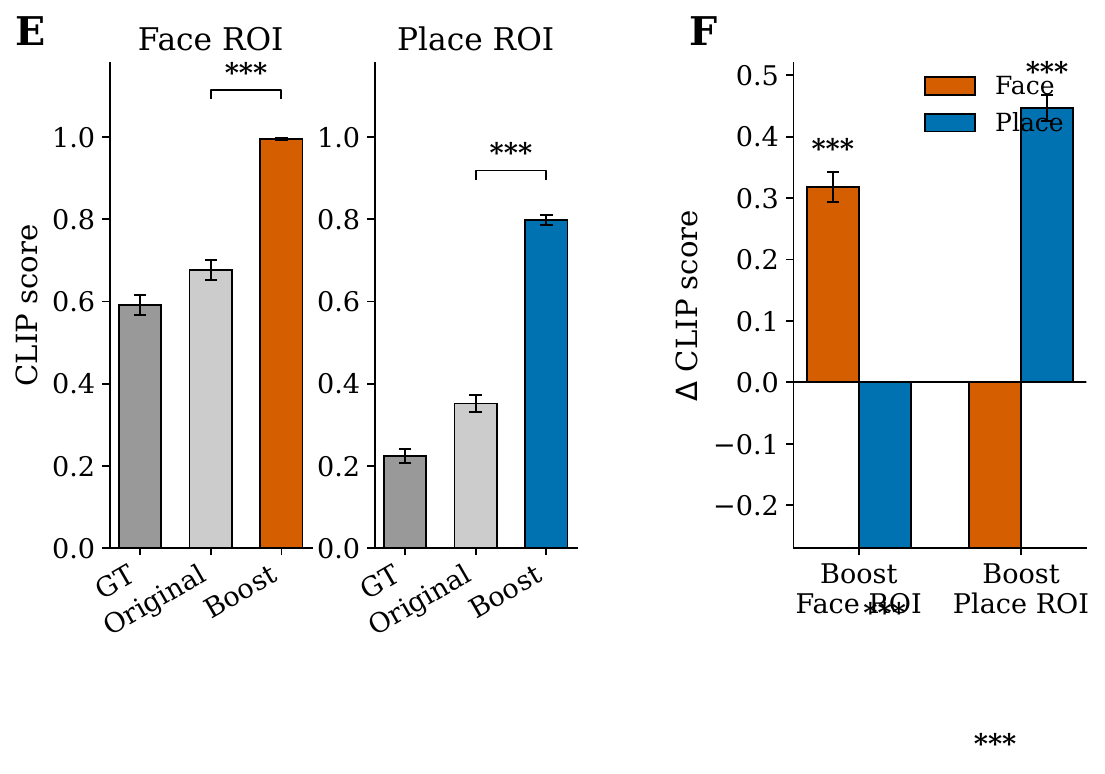}
    \caption{%
      \textbf{ROI-boost double-dissociation summaries for the HCP subjects.}
      Panels A--B show HCP~S1, panels C--D show HCP~S2, and panels E--F show HCP~S3. For each subject, the left panel reports the matched-category lift after boosting the corresponding ROI, and the right panel reports the cross-suppressive $\Delta$-score bars. These panels reproduce the per-subject analogue of main-text Fig.~\ref{fig:boost_and_dissociation}.%
    }
  \label{fig:app_dd_hcp}
\end{figure}

For completeness, we provide the per-subject visualizations for the five
additional subjects used in the supplementary ROI analyses. These include the
ROI-ablation stimulus-specificity panels, corresponding to the per-subject
analogue of main-text Fig.~\ref{fig:stimulus_specificity}, and the ROI-boost
double-dissociation summaries, corresponding to the per-subject analogue of
main-text Fig.~\ref{fig:boost_and_dissociation}. The stimulus-specificity
results are shown separately for the additional CC2017 subjects in
Fig.~\ref{fig:app_spec_cc2017} and for the HCP subjects in
Fig.~\ref{fig:app_spec_hcp}. The corresponding double-dissociation summaries
are shown in Figs.~\ref{fig:app_dd_cc2017} and~\ref{fig:app_dd_hcp},
respectively. The numerical values underlying these figures are reported in
Tables~\ref{tab:app_ablation} and~\ref{tab:app_enhance}.

\end{document}